\documentclass[10pt,twocolumn,letterpaper]{article}

\usepackage[pagenumbers]{cvpr} 

\usepackage{latexcolors}
\usepackage{indentfirst}
\usepackage{eqparbox}

\newcommand{\ctext}[3][RGB]{%
  \begingroup
  \definecolor{hlcolor}{#1}{#2}\sethlcolor{hlcolor}%
  \hl{#3}%
  \endgroup
}

\newcommand{\timetablefaster}[1]{{\colorbox[RGB]{208,240,192}{#1}}}
\newcommand{\timetablefastest}[1]{{\colorbox[RGB]{134,230,85}{#1}}}

\newcommand{\metrictablebest}[1]{{\colorbox[RGB]{134,230,85}{#1}}}
\newcommand{\metrictablesecond}[1]{{\colorbox[RGB]{208,240,192}{#1}}}

\definecolor{cvprblue}{rgb}{0.21,0.49,0.74}
\usepackage[pagebackref,breaklinks,colorlinks,allcolors=cvprblue]{hyperref}
\usepackage{multirow}        
\newcommand{\Mat}{\boldsymbol}
\newcommand{\Set}{\mathcal}

\newcommand{\real}{\mathbb{R}}

\DeclareMathOperator*{\argmin}{arg\,min}
\DeclareMathOperator*{\argmax}{arg\,max}

\usepackage{soul,xcolor} 
\usepackage{algorithm}
\usepackage{algorithmic}
\usepackage{placeins}
\usepackage[accsupp]{axessibility}

\def\acronym{Speedy-Splat}

\title{\acronym{}: Fast 3D Gaussian Splatting with \\
Sparse Pixels and Sparse Primitives}

\author{
Alex Hanson \hspace{2em} Allen Tu \hspace{2em} Geng Lin \hspace{2em} Vasu Singla \\
Matthias Zwicker \hspace{4em} Tom Goldstein \\
[0.65em]
\textnormal{University of Maryland, College Park}\\
[0.75em]
\url{https://speedysplat.github.io}\\
[0.75em]
}

\begin{document}
\setstcolor{red}

\twocolumn[{
\renewcommand\twocolumn[1][]{#1}
\maketitle
\begin{center}
    \vspace{-2.7mm} 
    \vspace{-20pt}
    \includegraphics[width=\linewidth]{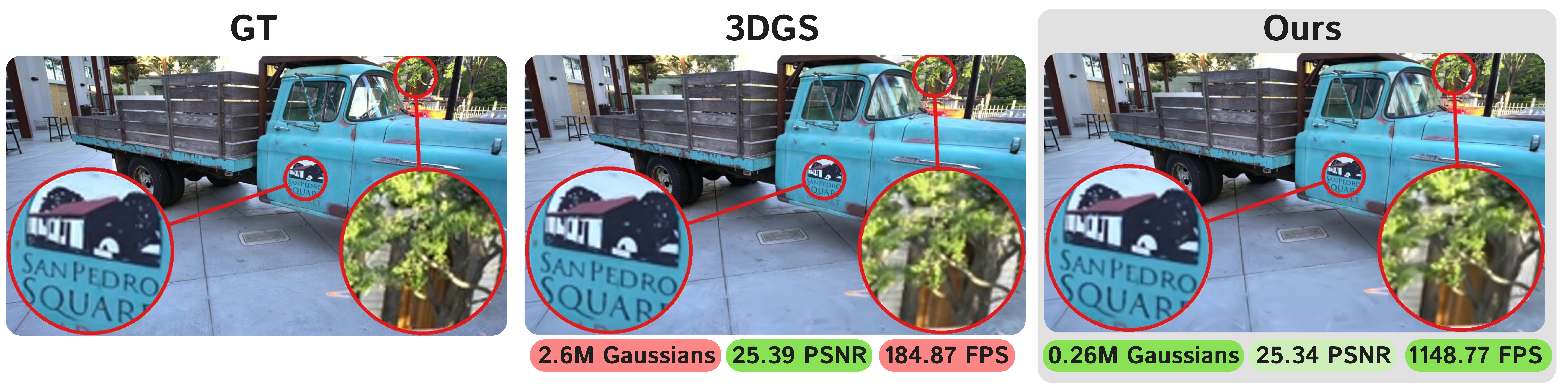}
    \vspace{-20pt}
    \captionsetup{type=figure}
    \caption{We reduce the number of Gaussians by over $90\%$, only marginally decrease PSNR, and accelerate rendering speed by $6.2\times$ in the Tanks \& Temples \emph{truck} scene when compared to 3D Gaussian Splatting (3D-GS). Additionally, we speed up training time by $1.38\times$.
    }
    \vspace{0mm}
    \label{fig:teaser}
\end{center}
}]

\begin{abstract}

3D Gaussian Splatting (3D-GS) is a recent 3D scene reconstruction technique that enables real-time rendering of novel views by modeling scenes as parametric point clouds of differentiable 3D Gaussians.
However, its rendering speed and model size still present bottlenecks, especially in resource-constrained settings.
In this paper, we identify and address two key inefficiencies in 3D-GS to substantially improve rendering speed. These improvements also yield the ancillary benefits of reduced model size and training time.
First, we optimize the rendering pipeline to precisely localize Gaussians in the scene, boosting rendering speed without altering visual fidelity.
Second, we introduce a novel pruning technique and integrate it into the training pipeline, significantly reducing model size and training time while further raising rendering speed.
Our \acronym{} approach combines these techniques to accelerate average rendering speed by a drastic $\mathit{6.71\times}$ across scenes from the Mip-NeRF 360, Tanks \& Temples, and Deep Blending datasets. 

\end{abstract}

\vspace{-2mm}
\section{Introduction}
\label{sec:intro}

Fast rendering of photorealistic novel views has been a long-standing goal in computer vision and graphics. Neural Radiance Fields (NeRF)~\cite{srinivasan2020nerf} and its variants have made significant strides in photorealistic 3D scene reconstruction by representing scenes as continuous neural volumetric models that encode scene density and color at spatial coordinates. However, despite recent efforts~\cite{mueller2022instant,Chen2022ECCVTensorf,Rivas-Manzaneque_2023_CVPRNerflight}, fast rendering in NeRF remains challenging because the volumetric sampling used in ray-marching is computationally expensive. Recently, 3D Gaussian Splatting (3D-GS)~\cite{kerbl3Dgaussians} has emerged as a promising alternative that enables real-time rendering by modeling scenes as parametric point clouds of differentiable 3D Gaussians. Nevertheless, its rendering speed is still limited by high parameter counts and certain algorithmic inefficiencies.
Efficient rendering is essential in applications such as virtual reality, networked systems, and multi-view streaming. Real-time rendering on resource-constrained edge devices, such as mobile phones, has yet to be achieved~\cite{lin2024rtgs}.

Although recent works on compressing 3D-GS models ~\cite{fan2023lightgaussian,hansontu2024pup,fang2024mini,niemeyer2024radsplat} achieve some speed-ups by reducing the number of parameters, few approaches directly target rendering speed~\cite{feng2024flashgs, lin2024rtgs}. In this paper, we specifically address this gap by demonstrating that the rendering speed of 3D-GS models can be drastically increased while maintaining competitive image quality. Additionally, we show that our methods reduce training time and substantially decrease model size. 

We begin by observing that the cost of 3D-GS rendering is proportional to both the number of Gaussians in the scene and the number of pixels processed per Gaussian. Our approach optimizes both factors. First, we reduce the number of pixels that are processed for each Gaussian by efficiently and accurately localizing it in the rendered image. Second, we reduce the total number of Gaussians in the model through a novel approach that maintains rendering quality.

3D-GS implements tile rendering to localize Gaussians in the image plane by assigning each Gaussian to the tiles that it intersects. We find the existing algorithm to be overly conservative; in Section~\ref{sec:snugbox}, we address this by introducing our SnugBox algorithm to precisely localize Gaussians by computing a tight bounding box around their extent. Then, in Section ~\ref{sec:accutile}, we extend SnugBox with our AccuTile algorithm to identify exact tile
intersections. Both approaches are plug-and-play, lead to respective inference speed-ups of $1.82\times$ and $1.99\times$ on average, and do not change the 3D-GS renderings.

To reduce the total number of Gaussians while preserving visual fidelity, we extend an existing pruning method~\cite{hansontu2024pup} by reducing its memory requirement by $36\times$ and incorporating it into the 3D-GS training pipeline. During the densification stage of 3D-GS training, Gaussians are regularly replicated and pruned. In Section~\ref{sec:soft_pruning}, we augment the densification stage with our Soft Pruning method to prune $80\%$ of Gaussians at regular intervals. After the densification stage, our Hard Pruning method, described in Section~\ref{sec:hard_pruning}, prunes an additional $30\%$ of Gaussians at set intervals. 

Our \acronym{} approach integrates both techniques into the 3D-GS training pipeline. On average, rendering speed is accelerated by $6.71\times$, model size is reduced by $10.6\times$, and training speed is improved by $1.4\times$ across all evaluated scenes while maintaining high image quality.

In summary, we propose the following contributions:

\begin{enumerate}
    \item \textbf{SnugBox:} A precise algorithm for computing Gaussian-tile bounding box intersections.
    \item \textbf{AccuTile:} An extension of SnugBox for computing exact Gaussian-tile intersections. 
    \item \textbf{Soft Pruning:} An augmentation for pruning Gaussians during densification.
    \item \textbf{Hard Pruning:} An augmentation for pruning Gaussians post-densification.
\end{enumerate}
\vspace{0.5mm} 
\section{Related work}
\label{sec:relate}

The real-time rendering speed of 3D-GS~\cite{kerbl3Dgaussians} on desktop GPUs has inspired research focused on further accelerating both its training and inference in resource-constrained environments. In this section, we review related works that specifically target these performance improvements.

\subsection{Pruning}
A large portion of Gaussians in vanilla 3D-GS models are redundant \cite{niemeyer2024radsplat, fan2023lightgaussian, fang2024mini}, motivating a recent line of work focused on pruning Gaussians from 3D-GS models to boost rendering speed with minimal loss of visual fidelity \cite{lin2024rtgs, fan2023lightgaussian, fang2024mini, girish2023eagles, lee2023compact, ali2024elmgs, ali2024trimming, niemeyer2024radsplat, hansontu2024pup}. Nearly all approaches assign a significance score to each Gaussian that is used to rank and prune them. Several works compute the aggregated ray contribution for each Gaussian across all input images \cite{fan2023lightgaussian, niemeyer2024radsplat, lee2023compact, fang2024mini}, while others combine opacity with additional information, such as gradients per Gaussian, to calculate their pruning criterion \cite{liu2024compgs, ali2024elmgs, ali2024trimming}. \citet{papantonakis2024reducing} use resolution and scale-aware redundancy metrics, EAGLES \cite{girish2023eagles} calculates the total transmittance per Gaussian, and \citet{lin2024rtgs} accelerate inference speed through an efficiency-aware strategy and foveated rendering.

PUP 3D-GS \cite{hansontu2024pup} introduces a more  principled approach by deriving a Hessian for each Gaussian that represents its sensitivity to the reconstruction error. While PUP 3D-GS achieves state-of-the-art post-hoc pruning results, computing its sensitivity score incurs considerable storage requirements that limit its viability for use during training. Our pruning approach directly extends PUP 3D-GS by improving its memory efficiency by $36\times$.

\subsection{Other Methods}

In addition to pruning, several other strategies have been explored to enhance the rendering and training speed of 3D-GS. Mini-Splatting~\cite{fang2024mini} modifies its densification strategies and adds a simplification stage to constrain the number of Gaussians. 3DGS-MCMC~\cite{kheradmand20243d} models training dynamics as an MCMC process. Revisiting Densification~\cite{rota2024revising} introduces an error-based densification strategy. 3DGS-LM \cite{hollein20243dgs} replaces Adam~\cite{Kingma2014AdamAM} with a tailored Levenberg-Marquardt optimizer~\cite{Gavin2013TheLM}. Taming 3DGS~\cite{mallick2024taming} proposes a constructive optimization process that limits the number of Gaussians to a pre-defined threshold set by the user. DISTWAR~\cite{durvasula2023distwar} dives into the low-level implementation of GPU thread scheduling and optimizes atomic processing with a novel primitive.
StopThePop~\cite{radl2024stopthepop} introduces a precise tile intersection method, but we find that our approach is notably faster in Appendix~\ref{sec:stp-ablation}. FlashGS \cite{feng2024flashgs} implements a tile intersection method similar to StopThePop.
Several works reduce training time and memory requirements by enforcing geometric constraints~\cite{scaffoldgs,ren2024octree,wei2024normal,ververas2024sags}. Most of these approaches are orthogonal to ours and can be applied alongside it.

\section{Background}
\label{sec:background}

\subsection{3D Gaussian Splatting Overview}

\begin{table*}[t]
\caption{
Average execution time (milliseconds) of each function across all scenes in Section~\ref{sec:datasets}. The operation in each row is applied \textbf{cumulatively} to all of the following rows. For each model, accurate measurements are collected by averaging execution times across three runs that each render the test set 20 times to reduce variance. The \ctext[RGB]{134,230,85}{fastest} and \ctext[RGB]{208,240,192}{second} fastest times are color coded. }
\centering
\resizebox{\textwidth}{!}{
\begin{tabular}{l|cccccc|c}
\toprule
Method &
Preprocess &
Inclusive Sum &
Duplicate with Keys &
Radix Sort &
Identify Tile Ranges &
Render &
Overall \\
\midrule
Baseline &
0.665 & 0.045 & 0.570 & 1.551 & 0.082 & 4.483 & 7.478
\\
+SnugBox &
0.656 & 0.046 & 0.208 (2.738$\times$) & 0.729 (2.126$\times$) & 0.041 (1.980$\times$) & 2.344 (1.913$\times$) & 4.102 (1.823$\times$)
\\
+AccuTile &
0.668  & 0.046 & 0.221 (2.575$\times$) & 0.612 (2.533$\times$) & 0.035 (2.326$\times$) & 2.062 (2.175$\times$) & 3.748 (1.995$\times$)
\\
+Soft Pruning &
\timetablefaster{0.370 (1.798$\times$)} & \timetablefaster{0.030 (1.494$\times$)} & \timetablefaster{0.146 (3.906$\times$)} & \timetablefaster{0.404 (3.843$\times$)} & \timetablefaster{0.024 (3.422$\times$)} & \timetablefaster{1.337 (3.354$\times$)} & \timetablefaster{2.381 (3.141$\times$)}
\\
+Hard Pruning &
\timetablefastest{0.091 (7.293$\times$)} & \timetablefastest{0.016 (2.769$\times$)} & \timetablefastest{0.090 (6.325$\times$)} & \timetablefastest{0.215 (7.217$\times$)} & \timetablefastest{0.013 (6.537$\times$)} & \timetablefastest{0.619 (7.247$\times$)} & \timetablefastest{1.114 (6.712$\times$)} \\
\bottomrule
\end{tabular}
}
\label{tab:time_table_full}
\vspace{-2mm}
\end{table*}

3D Gaussian Splatting (3D-GS)~\cite{kerbl3Dgaussians} models scenes as parametric, point-based representations that use differentiable 3D Gaussians as primitives.
Given a set of ground truth training images $\Set{I}_{gt} = \{\Mat{I}_i \in \real^{H \times W}\}_{i=1}^K$, the scene is initialized by using Structure from Motion (SfM) to produce a sparse point cloud that serves as the initial means for the 3D Gaussians. The estimated camera poses $\Set{P}_{gt} = \{\phi_i \in \real^{3 \times 4}\}_{i=1}^K$ are paired with their corresponding images and are used to optimize the scene.

Each 3D Gaussian primitive $\Set{G}_i$ is parameterized by three geometry parameters -- mean $\mu_i \in \real^3$, scale $s_i \in \real^3$, and rotation $r_i \in \real^4$ -- and two color parameters -- view-dependent spherical harmonics $h_i \in \real^{16\times 3}$ and opacity $\sigma_i\in\real$.
The set of all parameters can be described as:
\begin{equation}
\Set{G} = \{\Set{G}_i = \{\mu_i, s_i, r_i, h_i, \sigma_i\}\}_{i=1}^N,
\end{equation}
where $N$ is the number of Gaussians in the model.

Given camera pose $\phi$, the scene is rendered by projecting all Gaussians to image space and applying alpha blending to each pixel. Models are optimized via stochastic gradient descent on image reconstruction losses:
\begin{equation}
    L(\Set{G} | \phi, I_{gt} ) = ||I_{\Set{G}}(\phi)-I_{gt}||_1 + L_{\text{D-SSIM}}(I_\Set{G}(\phi),I_{gt}),
    \label{eq:gs_loss}
\end{equation}
where $I_{\Set{G}(\phi)}$ is the rendered image for pose $\phi$.

During optimization, the scene is periodically densified by cloning and splitting uncertain Gaussians and pruned by removing large and transparent Gaussians. The opacities of the Gaussians are also periodically reset.

\vspace{0.5ex}
\subsection{3D Gaussian Splatting Rendering Specifics}

3D-GS uses a tile-based rendering strategy that divides the rendered image into $16 \times 16$ pixel tiles. Each Gaussian is projected into image space, where its intersection with these tiles is computed. 
Then, these Gaussian-to-tile mappings are sorted to collect and order the Gaussians by depth for pixel-wise rendering.

Rendering runtime is dominated by six key functions. Table \ref{tab:time_table_full} empirically analyzes the execution time of each function and highlights the improvements achieved by our methods. Descriptions of each function are provided in the following sections.

\vspace{1mm}
\subsubsection{Preprocessing}

The \texttt{preprocess} kernel is 
parallelized such that each thread processes a single Gaussian $\Set{G}_i$. It computes a 2D projection of $\Set{G}_i$ to image space and obtains a count of tiles that it intersects. 

The mean $\mu_i$ is projected to image space using a viewing transform $\Mat{W}$ and a perspective projection, yielding 2D mean $\mu_{i_{2D}}$ and depth that are stored for later processing. The scale $s_i$ and rotation $r_i$ parameters are converted to the diagonal scale $\Mat{S}_i$ and rotation $\Mat{R}_i$ matrices. The 3D covariance is then defined as:

\begin{equation}
\Mat{\Sigma}_{i_{3D}} = \Mat{R}_i\Mat{S}_i\Mat{S}_i^T\Mat{R}_i^T,
\end{equation}

which is projected via:

\begin{equation}
\Mat{\hat{\Sigma}_{i_{3D}}} = \Mat{J} \Mat{W} \Mat{\Sigma_{i_{3D}}} \Mat{W}^T \Mat{J}^T,
\end{equation}

where $\Mat{J}$ is the Jacobian of the first order approximation of the perspective projection. Dropping the last row and column of $\Mat{\hat{\Sigma}_{i_{3D}}}$ gives $\Mat{\Sigma_{i_{2D}}}$~\cite{kerbl3Dgaussians,zwicker2002ewa}. The largest eigenvalue of $\Mat{\Sigma_{i_{2D}}}$ is used to compute the count of tiles intersected by this Gaussian $\Set{G}_i$ as shown in Figure~\ref{fig:3dgs_tile_intersect}.
$\Mat{\Sigma_{i_{2D}}}^{-1}$ is computed, along with a view-dependent color $c_i$, derived from $W$ and $h_i$. All three values are stored for later processing.

\subsubsection{Sorting}
After preprocessing, four functions process the Gaussians for pixel-wise rendering:
\begin{itemize}
    \item \textbf{\texttt{InclusiveSum}:} A \texttt{CUDA} primitive that computes the prefix sum over all Gaussian-tile counts to allocate key and value arrays for Gaussian-to-tile mapping.
    \item \textbf{\texttt{duplicateWithKeys}:} A Gaussian-parallel kernel that recomputes Gaussian-tile intersections to generate a key for each intersecting tile, consisting of the tile index and Gaussian depth.
    \item \textbf{\texttt{RadixSort}: } A \texttt{CUDA} primitive that sorts the key array, ordering Gaussian indices by tile and then depth.
    \item \textbf{\texttt{identifyTileRanges}:} A kernel parallelized across the key array to post-process keys before pixel rendering.
\end{itemize}

\subsubsection{Rendering}
The \texttt{render} kernel is parallelized across pixels. For each pixel $p$, all Gaussians within its corresponding tile are loaded and processed in depth order as determined by \texttt{RadixSort}. An alpha value:

\begin{equation}
\alpha_i(p) = \sigma_i g_i(p)
\label{eq:render_alpha}
\end{equation}

is computed for each Gaussian $\Set{G}_i$, where $g_i$ is the value of the projected 2D Gaussian at pixel $p$: 
\begin{equation}
g_i=e^q, q = -\frac{1}{2} (p - \mu_{i_{2D}}) \Mat{\Sigma_{i_{2D}}}^{-1} (p - \mu_{i_{2D}})^T.
\label{eq:2d_gaussian_scalar}
\end{equation}

If $\alpha_i > \frac{1}{255}$, then the Gaussian is included in the alpha compositing of the pixel color $C$, given by:

\begin{equation}
    C(p)=\sum_{i\in \Set{N}} c_i \alpha_i(p) \prod_{j=1}^{i-1}(1 - \alpha_j(p)).
    \label{contribution}
\end{equation}

\section{Methods}
\label{sec:method}

Our \acronym{} methods are motivated by two key insights into inefficiencies within the 3D-GS rendering pipeline. First, Gaussian Splatting grossly overestimates the extent of Gaussians in the image. Second, as demonstrated by ~\cite{hansontu2024pup} and other recent pruning works, 3D-GS models are heavily overparameterized.

\subsection{Precise Tile Intersect}
\label{sec:tile_intersect}
Gaussian Splatting identifies tiles intersected by Gaussian $\Set{G}_i$ by calculating the maximum eigenvalue $\lambda_{\max}$ of its projected 2D covariance $\Mat{\Sigma_{i_{2D}}}$, then selecting all tiles that intersect the square inscribing the circle defined by center $\mu_{i_{2D}}$ and radius:

\begin{equation}
r = \left\lceil 3 \sqrt{\lambda_{\max}} \right\rceil.
\label{eq:radius}
\end{equation}

This approach neglects opacity $\sigma_i$ in its calculation and generally overestimates the Gaussian extent, as illustrated by Figure~\ref{fig:3dgs_tile_intersect}. The actual extent of Gaussian $\Set{G}_i$, shown in Figure~\ref{fig:snugbox_tile_intersect}, is given by the threshold placed on its alpha value $\alpha_i$. Specifically, $\Set{G}_i$ does not contribute to the rendering of pixel $p$ if $\alpha_i < \frac{1}{255}$ for $p$. By applying the actual extent in tile intersection calculations, we arrive at a far more concise set of intersected tiles.

We now show the derivation of this extent. The furthest pixel extent of Gaussian $\Set{G}_i$ can be determined by directly substituting this threshold into Equation~\ref{eq:render_alpha}. Rearranging terms gives:
\begin{equation}
2 \log(255 \sigma_i) = (p - \mu_{i_{2D}}) \Mat{\Sigma_{i_{2D}}}^{-1} (p - \mu_{i_{2D}})^T.
\end{equation}
We can rewrite:
\begin{equation}
p =  \begin{pmatrix} p_x \\ p_y \end{pmatrix}, 
\mu_{i_{2D}} = \begin{pmatrix} \mu_x \\ \mu_y \end{pmatrix}, 
\Mat{\Sigma_{i_{2D}}}^{-1} = \begin{pmatrix} a & b \\ b & c \end{pmatrix}.
\end{equation}
Specifying threshold $t$ and centered coordinates $x_d$ and $y_d$:
\begin{gather}
t = 2\log(255\sigma_i), \\
x_d = p_x - \mu_x, \\
y_d = p_y - \mu_y,
\end{gather}
gives the pixel extent with coordinates $x_d$ and $y_d$ that satisfy the ellipse equation: 
\begin{equation}
t = a x_d^2 + 2 b x_d y_d + c y_d^2.
\label{eq:simplified_ellipse}
\end{equation}

Our approach uses this pixel extent to reduce the number of tiles contained in the Gaussian-to-tile mappings for each Gaussian. 
We propose two methods for computing precise tile intersections. 
First, our \textbf{SnugBox} algorithm produces a tight bounding box around each Gaussian. 
Second, our \textbf{AccuTile} algorithm extends it to identify the exact set of tiles intersected by the Gaussian.

\subsubsection{SnugBox}
\label{sec:snugbox}

Our SnugBox method uses this elliptical extent to compute an axis-aligned bounding box that more precisely identifies tiles intersected by Gaussian $\Set{G}_i$. 
To derive this bounding box, we rearrange Equation~\ref{eq:simplified_ellipse} to solve for $y_d$:
\begin{equation}
y_d = \frac{-b x_d \pm \sqrt{(b^2 - a c) x_d^2 + t c}}{c}.
\label{eq:ellipse_intersect}
\end{equation}
To find the $y$-coordinate bounding box edges $y_{\min}$ and $y_{\max}$, we identify the values of $x_d$ where $\partial y_d / \partial x_d = 0$. We refer to these $x_d$ values as $x_{d_{args}}$ to specify that they are the $\argmin y_d$ and $\argmax y_d$ values. Differentiating Equation~\ref{eq:ellipse_intersect} and solving for $x_{d_{args}}$ gives:

\begin{equation}
x_{d_{args}} = \pm \sqrt{\frac{-b^2 t }{(b^2 - ac) a}}.
\label{eq:bbox_args}
\end{equation}

Substituting $x_{d_{args}}$ into Equation~\ref{eq:ellipse_intersect} and adding $\mu_y$ gives $y_{\min}$ and $y_{\max}$. Due to the symmetry of Equation~\ref{eq:simplified_ellipse}, we can find the $x$-coordinate bounding box edges $x_{min}$ and $x_{max}$ by swapping $y_d$ and $x_d$ and constants $a$ and $c$ to rewrite Equations~\ref{eq:ellipse_intersect} and ~\ref{eq:bbox_args} in terms of $x_d$ and $y_{d_{args}}$.

\begin{figure*}[t]
    \centering
    \begin{subfigure}[b]{0.32\textwidth}
        \centering
        \includegraphics[width=\textwidth]{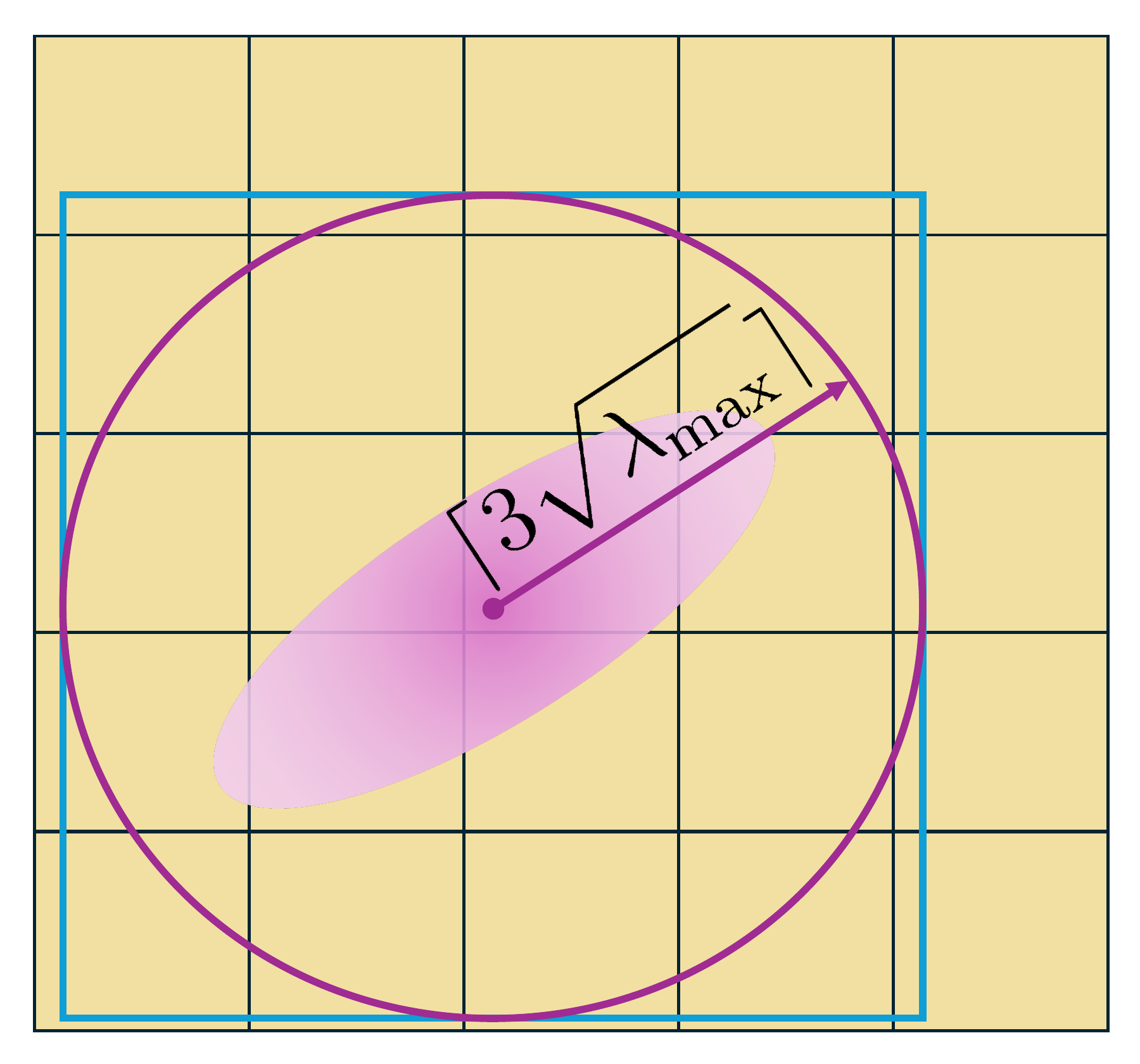}
        \caption{3D Gaussian Splatting}
        \label{fig:3dgs_tile_intersect}
    \end{subfigure}
    \begin{subfigure}[b]{0.32\textwidth}
        \centering
        \includegraphics[width=\textwidth]{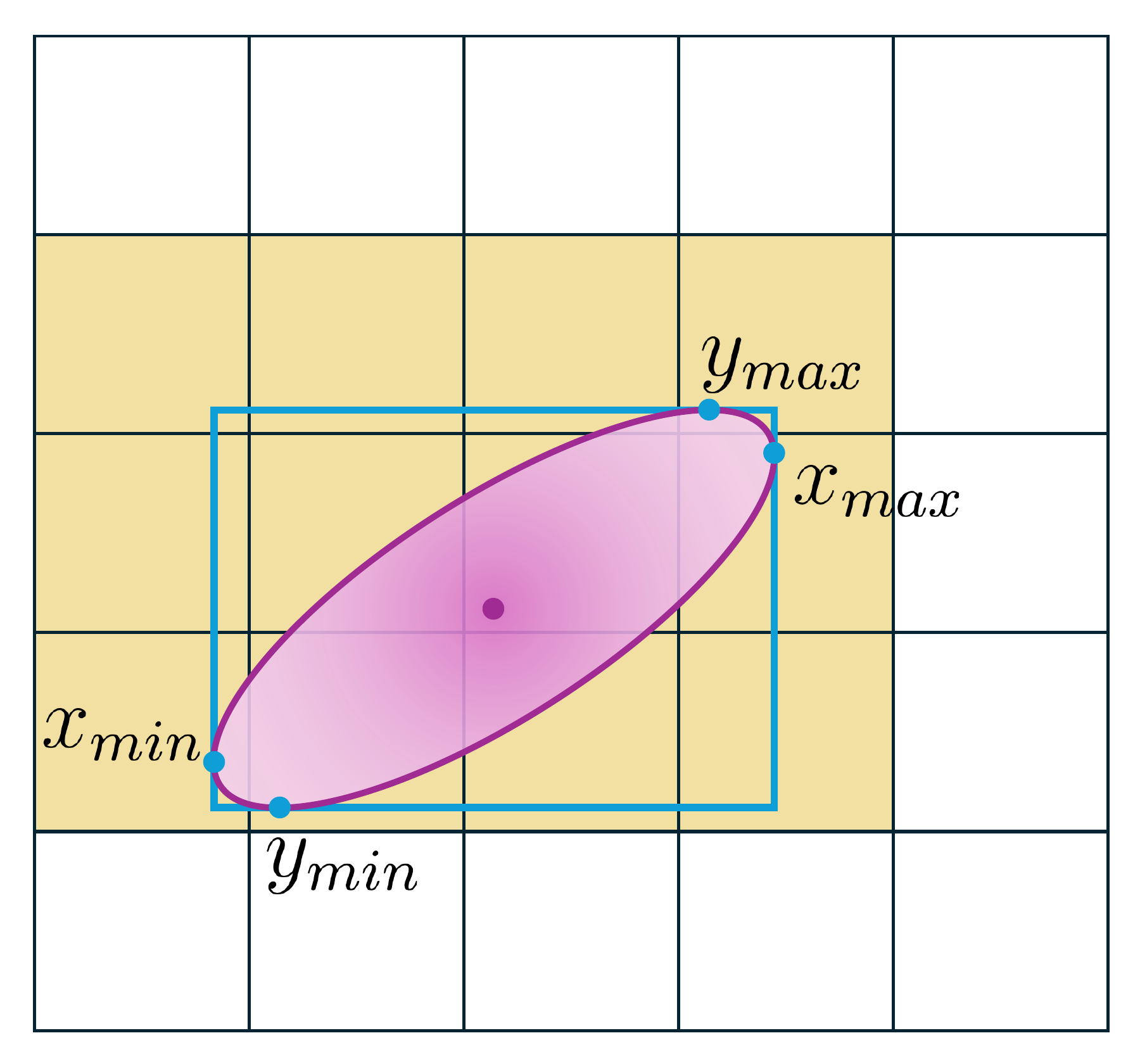} 
        \caption{SnugBox}
        \label{fig:snugbox_tile_intersect}
    \end{subfigure}
    \begin{subfigure}[b]{0.32\textwidth}
        \centering
        \includegraphics[width=\textwidth]{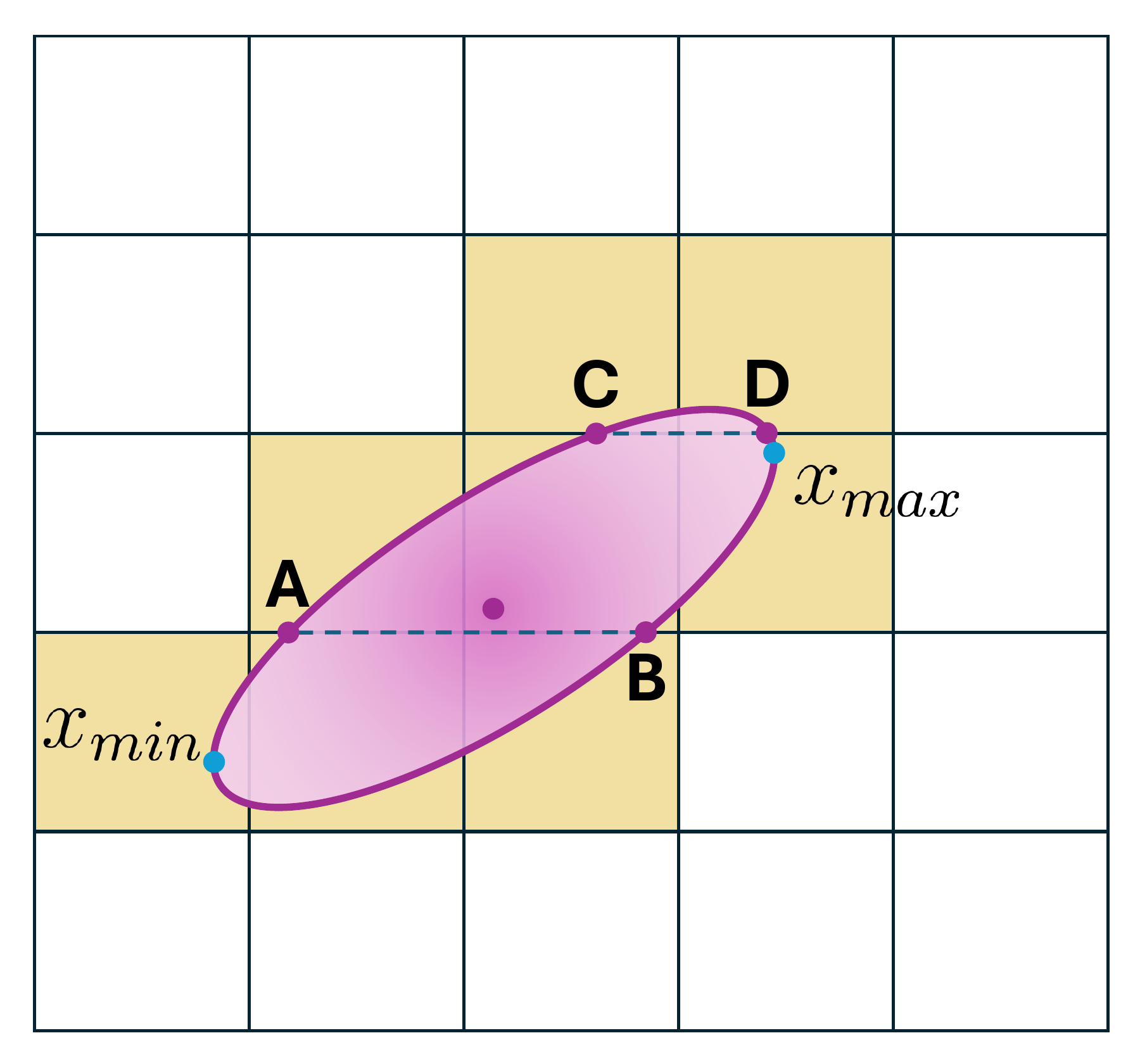}
        \caption{AccuTile}
        \label{fig:accutile_tile_intersect}
    \end{subfigure}
    \caption{\textbf{Gaussian tile allocation by method.}
    (a) 3D Gaussian Splatting allocates a Gaussian to a tile when that tile \ctext[RGB]{240,237,199}{\textbf{intersects}} the \ctext[RGB]{132,216,255}{\textbf{square}} inscribing the \ctext[RGB]{223,166,220}{\textbf{circle}} with radius $\lceil 3 \sqrt{\lambda_{\max}} \rceil$ defined in Equation~\ref{eq:radius}.
    (b) Our SnugBox method allocates a Gaussian to a tile when that tile \ctext[RGB]{240,237,199}{\textbf{intersects}} the tight \ctext[RGB]{132,216,255}{\textbf{bounding box}} defined by the axis-aligned minima and maxima of the \ctext[RGB]{223,166,220}{\textbf{ellipse}} given by Equation~\ref{eq:simplified_ellipse}.
    (c) Our AccuTile method allocates a Gaussian to a tile only if that tile \ctext[RGB]{240,237,199}{\textbf{intersects}} the \ctext[RGB]{223,166,220}{\textbf{ellipse}} via Algorithm~\ref{alg:accutile}, which computes the minimum and maximum tiles by iterating over the shorter side of the rectangular tile extent given by SnugBox. In this example, our AccuTile algorithm iterates over the tile rows; the only points that are processed are \textbf{$x_{min}$}, \textbf{$x_{max}$}, \textbf{A}, \textbf{B}, \textbf{C}, and \textbf{D}.
    }
    \label{fig:all_tile_intersect}
    \vspace{-2mm}
\end{figure*}

After the bounding box edges are identified, our method follows 3D-GS and converts these edges to tile indices by dividing by tile size, rounding, and clipping to the image boundary. As depicted in Figure~\ref{fig:all_tile_intersect}, Snugbox can produce a significantly tighter bounding box than 3D-GS.
Meanwhile, its computational overhead is small as it performs a constant number of operations and is only called twice in the rendering pipeline -- once to count the Gaussians intersecting tiles in \texttt{preprocess} and once to populate the Gaussian-to-tile arrays in \texttt{duplicateWithkeys}. Table~\ref{tab:time_table_full} reports that SnugBox improves the efficiency of all downstream functions and produces an average overall speed-up of $1.82\times$.

\subsubsection{AccuTile}
\label{sec:accutile}

\begin{algorithm}[t]
\caption{The AccuTile Algorithm. For simplicity, the algorithm outlined here is applied to the rows of the SnugBox tile extent bounding box, matching the example in Figure~\ref{fig:accutile_tile_intersect}. In practice, it is applied along the smaller side of the tile extent. The subscripts $t$, $b$, $l$, and $r$ represent the \emph{top}, \emph{bottom}, \emph{left}, and \emph{right} sides, respectively. A proof of correctness sketch is presented in Appendix~\ref{sec:appendix:proof}.}
\label{alg:accutile}
\vspace{2mm}
\begin{algorithmic}
{
    \newcommand{\ellipseE}{\colorbox[RGB]{223,166,220}{\textbf{E}}}
    
    \newcommand{\snugBoxB}{\colorbox[RGB]{132,216,255}{\textbf{B}}}
    \newcommand{\snugBoxBb}{\colorbox[RGB]{132,216,255}{\textbf{B$_b$}}}
    \newcommand{\snugBoxBt}{\colorbox[RGB]{132,216,255}{\textbf{B$_t$}}}
    \newcommand{\snugBoxBl}{\colorbox[RGB]{132,216,255}{\textbf{B$_l$}}}
    \newcommand{\snugBoxBr}{\colorbox[RGB]{132,216,255}{\textbf{B$_r$}}}
    
    \newcommand{\tileRect}{\colorbox[RGB]{240,237,199}{\textbf{R}}}
    \newcommand{\tileRectb}{\colorbox[RGB]{240,237,199}{\textbf{R$_b$}}}
    
    \newcommand{\intersectionIMin}{\colorbox[RGB]{223,166,220}{\textbf{i$_{\min}$}}}
    \newcommand{\intersectionIMax}{\colorbox[RGB]{223,166,220}{\textbf{i$_{\max}$}}}
    
    \newcommand{\ellipseRowMin}{\colorbox[RGB]{223,166,220}{e$_{\min}$}}
    \newcommand{\ellipseRowMax}{\colorbox[RGB]{223,166,220}{e$_{\max}$}}
    
    \newcommand{\lineMin}{\colorbox[RGB]{240,237,199}{line$_{\min}$}}
    \newcommand{\lineMax}{\colorbox[RGB]{240,237,199}{line$_{\max}$}}
    
    \newcommand{\tileRowMin}{\colorbox[RGB]{240,237,199}{tile$_{\min}$}}
    \newcommand{\tileRowMax}{\colorbox[RGB]{240,237,199}{tile$_{\max}$}}
    
    \newcommand{\tileRow}{\colorbox[RGB]{240,237,199}{r}}
    \newcommand{\tileRowtop}{\colorbox[RGB]{240,237,199}{r$_t$}}
    
    \newcommand{\tileCountC}{\colorbox[RGB]{240,237,199}{C}}

    \REQUIRE Ellipse \ellipseE \COMMENT{Eq.~\ref{eq:simplified_ellipse}}
    \REQUIRE SnugBox Bounding Box \snugBoxB
    \REQUIRE SnugBox Tile Extent Rectangle \tileRect
    
    \STATE \hspace{-1em} \textbf{Initialize:} Tile count \tileCountC $\gets 0$

    \STATE \lineMin $\gets$ \tileRectb

    \IF { \lineMin $\ge$ \snugBoxBb}
        \STATE{ \intersectionIMin $\gets $ Intersections(\lineMin, \ellipseE)} \COMMENT{Eq.~\ref{eq:ellipse_intersect}}
    \ENDIF

    \FOR{row \tileRow \textbf{ in} \tileRect}
        
        \STATE \lineMax $\gets$ \tileRowtop

        \IF { \lineMax $\le$ \snugBoxBt}
            \STATE{ \intersectionIMax $\gets $ Intersections(\lineMax, \ellipseE)} \COMMENT{Eq.~\ref{eq:ellipse_intersect}}
        \ENDIF
        
        \STATE \ellipseRowMin $\gets$ \snugBoxBl  \textbf{ if} \snugBoxBl \textbf{ in} \tileRow \textbf{ else} min(\intersectionIMin, \intersectionIMax
        )
        
        \STATE \ellipseRowMax $\gets$ \snugBoxBr \textbf{ if} \snugBoxBr \textbf{ in} \tileRow \textbf{ else} max(\intersectionIMin, \intersectionIMax)

        \STATE \tileRowMin, \tileRowMax $\gets$ Convert(\ellipseRowMin, \ellipseRowMax)
        
        \STATE \tileCountC $\gets$ \tileCountC + (\tileRowMax - \tileRowMin)
        
        \STATE Process(\tileRowMin, \tileRowMax)
        
        \STATE \intersectionIMin $\gets$ \intersectionIMax
    \ENDFOR
    \STATE \textbf{return} \tileCountC 
}
\end{algorithmic}
\vspace{2mm}
\end{algorithm}

Our AccuTile method, outlined in Algorithm~\ref{alg:accutile}, extends SnugBox to identify the exact tiles intersected by the Gaussian. It takes as input the tight bounding box produced by Snugbox and its rectangular tile extent -- depicted as the blue box and yellow tiles, respectively, in Figure~\ref{fig:snugbox_tile_intersect}. Depending on which dimension is smaller, AccuTile then processes either the rows or columns of this tile extent to determine the exact tiles that intersect the Gaussian. Specifically, we identify the minimum and maximum extent of the ellipse within a given row or column, then convert those points to the corresponding touched tiles. All tiles between the minimum and maximum tiles intersect the ellipse.

AccuTile's key insight is that calculating the minimum and maximum extent of the ellipse within each row or column requires computing only two points per iteration. The only possible inflection points of the elliptical curve are the bounding box minimum and maximum points identified by Snugbox. If one or both of these points lie within the tile row or column, then they represent the minimum or maximum extent of the ellipse there. If neither point is within the row or column, then the points along that boundary side are monotonically decreasing or increasing -- the minimum or maximum point must lie on one of the boundary lines of that row or column. Since the boundary of the last row or column is shared with the next one, we only need to compute the intersection of the ellipse with the next boundary line in each iteration. Thus, our AccuTile algorithm counts tiles in time proportional to the shorter side of the tile extent and processes tiles in time proportional to the tile count.

Figure~\ref{fig:all_tile_intersect} illustrates how Accutile restricts the tight \ctext[RGB]{132,216,255}{bounding box} produced by SnugBox to the exact tiles that the Gaussian touches. The \ctext[RGB]{240,237,199}{tile extent rows} from Snugbox, shown in Figure~\ref{fig:snugbox_tile_intersect}, are processed starting from the bottom. The \ctext[RGB]{240,237,199}{lower tile row boundary} falls below the \ctext[RGB]{132,216,255}{bounding box}, so no initial intersection is computed. The \ctext[RGB]{240,237,199}{upper boundary} lies below the top of the \ctext[RGB]{132,216,255}{bounding box}, so intersection points \ctext[RGB]{223,166,220}{$A$} and \ctext[RGB]{223,166,220}{$B$} are calculated using Equation~\ref{eq:ellipse_intersect}. \ctext[RGB]{132,216,255}{$x_{\min}$} is  assigned as this row's minimum ellipse extent \ctext[RGB]{223,166,220}{$e_{\min}$} because it lies within it, and \ctext[RGB]{223,166,220}{$B$} is assigned as its maximum ellipse extent \ctext[RGB]{223,166,220}{$e_{\max}$} because it is the maximal point. Consequently, this row's \ctext[RGB]{240,237,199}{tile extent} is from the tile containing \ctext[RGB]{132,216,255}{$x_{\min}$} to the tile containing \ctext[RGB]{223,166,220}{$B$}. For the next row, we keep points \ctext[RGB]{223,166,220}{$A$} and \ctext[RGB]{223,166,220}{$B$} and compute \ctext[RGB]{240,237,199}{upper boundary} points \ctext[RGB]{223,166,220}{$C$} and \ctext[RGB]{223,166,220}{$D$}. \ctext[RGB]{223,166,220}{$A$} and \ctext[RGB]{132,216,255}{$x_{\max}$} are assigned as the row's \ctext[RGB]{223,166,220}{$e_{\min}$} and \ctext[RGB]{223,166,220}{$e_{\max}$}, and the tiles containing them are its \ctext[RGB]{240,237,199}{tile extent}. Finally, for the last row, \ctext[RGB]{223,166,220}{$C$} and \ctext[RGB]{223,166,220}{$D$} are kept; no additional points are computed because the row's \ctext[RGB]{240,237,199}{upper boundary} is above the \ctext[RGB]{132,216,255}{bounding box}. Thus, the \ctext[RGB]{240,237,199}{tile extent} of this row is between \ctext[RGB]{223,166,220}{$C$} and \ctext[RGB]{223,166,220}{$D$}.

The number of tiles is further reduced from Figure~\ref{fig:snugbox_tile_intersect} to  Figure~\ref{fig:accutile_tile_intersect}, much less than the original 3D-GS method in Figure~\ref{fig:3dgs_tile_intersect}. In Table~\ref{tab:time_table_full}, we report that AccuTile further accelerates all downstream functions, culminating in an average overall speed-up of $1.99\times$.

\subsection{Efficient Pruning}
\label{sec:efficient_pruning}

PUP 3D-GS~\cite{hansontu2024pup} is a pruning method that quantifies the sensitivity of each Gaussian to training views, removing a set percentage with the lowest sensitivities. Sensitivity is computed by approximating the Hessian of the $L_2$ loss:
\begin{equation}
H = \nabla_{\Set{G}}^2 L_2 = \sum_{\phi \in \Set{P}_{gt}} \nabla_{\Set{G}} I_{\Set{G}}(\phi) \nabla_{\Set{G}} I_{\Set{G}}(\phi)^T, 
\end{equation}
where $\nabla_{\Set{G}} I_{\Set{G}}(\phi)$ is the gradient over all Gaussian parameters on the rendered image $I_{\Set{G}}$ for pose $\phi$. $H$ is shown to be exact when the $L_1$ residual error vanishes~\cite{hansontu2024pup}.

A per-Gaussian sensitivity can be derived by splitting $H$ into the block diagonal elements that only capture inter-Gaussian parameter relationships:
\begin{equation}
H_i = \sum_{\phi \in \Set{P}_{gt}} \nabla_{\Set{G}_i} I_{\Set{G}}(\phi) \nabla_{\Set{G}_i} I_{\Set{G}}(\phi)^T,
\end{equation}
where $\nabla_{\Set{G}_i}$ is the gradient with respect to only $\Set{G}_i$. This measures the sensitivity of the $L_2$ loss with respect to Gaussian $\Set{G}_i$, assuming all other Gaussians are held constant.

$H_i$ is again approximated by only using the six mean $\mu_i$ and scale $s_i$ parameters to specifically capture geometric sensitivity. The log determinant is taken to provide a representative scalar score $U_i$:
\begin{equation}
U_i = \log | \nabla_{\mu_i,s_i} I_{\Set{G}} \nabla_{\mu_i,s_i} I_{\Set{G}}^T |.
\label{eq:pup_score}
\end{equation}

Using this score, up to $90\%$ of Gaussians can be robustly pruned from the model while retaining high visual quality. 

Although PUP 3D-GS touts high compression ratios and rendering speeds, we identify two key drawbacks in its formulation. First, computing the Hessian requires storage proportional to $N\times36$, where $N$ is the number of Gaussians. In comparison, the 3D-GS model has a memory footprint proportional to $N\times59$ because it stores $59$ parameters per Gaussian. While this score is effective for post-hoc pruning, using it during training is impractical. 

Second, computing the Hessian requires the pixel-wise gradients of $\mu$ and $s$. Since these are 3D parameters of the Gaussian primitives, obtaining their gradients requires back-propagating through the \texttt{render} kernel parallelized per pixel, then back-propagating each Gaussian contributing to that pixel in its thread. This breaks the efficient flow of gradients in 3D-GS, where the per-pixel gradients from the \texttt{render} kernel are parallelized and aggregated to the 2D $\mu_{2D}$ and $\Sigma_{2D}$ parameters, which are then parallelized across Gaussians to compute gradients for $\mu$ and $s$. 

Our approach builds on PUP 3D-GS by introducing an \textbf{efficient pruning score} that we incorporate into the 3D-GS training pipeline. We also define two distinct pruning modalities: \textbf{Soft Pruning}, which takes place during densification in the first $15000$ iterations, and \textbf{Hard Pruning}, which is applied after densification is completed after iteration $15000$.

\subsubsection{Efficient Pruning Score}

Our insight is that both drawbacks can be alleviated by reparameterizing the Hessian. Concretely, we express the influence of all spatial parameters of Gaussian $\Set{G}_i$ by computing the Hessian approximation with respect to the 2D projected value of $\Set{G}_i$ at pixel $p$, given by $g_i(p)$ in Equation~\ref{eq:2d_gaussian_scalar}. In doing so, the pruning score $U_i$ from Equation~\ref{eq:pup_score} can be expressed as:

\begin{equation}
\tilde{U}_i = \log | \nabla_{g_i} I_{\Set{G}} \nabla_{g_i} I_{\Set{G}}^T |.
\end{equation}
Since $g_i$ is a scalar and $\log$ is monotonically increasing, we can rewrite this score as:
\begin{equation}
\tilde{U}_i = (\nabla_{g_i} I_{\Set{G}})^2.
\end{equation}

Gradient $\nabla_{g_i} I_{\Set{G}}$ is already computed in the backward pass of \texttt{render} and can be efficiently squared and aggregated across all pixels. Moreover, the maximum space requirement for this score is proportional to the number of Gaussians $N$, reducing the storage requirement by $36\times$ and allowing this score to be used during training. 

\subsubsection{Soft Pruning}
\label{sec:soft_pruning}

To maintain a robust Hessian approximation, we observe that the $L_1$ loss becomes quite small by iteration $6000$ and remains low except after an opacity reset is performed. As such, we augment the densification pipeline to include our Soft Pruning method, where the model is pruned immediately before the three opacity resets at $6000$, $9000$, and $12000$ iterations. Surprisingly, we find that we can set extremely high Soft Pruning ratios -- in our experiments, visual fidelity is preserved at $80\%$ pruning.

\begin{figure*}[t]
  \includegraphics[width=\linewidth]{figures/speedy_splat_comparison.pdf}
  \caption{Visual comparison of 3D-GS, PUP 3D-GS, and our method. Notice that, while reaching similar compression ratios to PUP 3D-GS, our \acronym{} method delivers vastly faster rendering speeds. Top: \emph{playroom} from the Deep Blending dataset. Middle: \emph{bicycle} from the Mip-NeRF 360 dataset. Bottom: \emph{drjohnson} from the Deep Blending dataset.}
  \label{fig:experiments}
  \vspace{-2mm}
\end{figure*}

\subsubsection{Hard Pruning}
\label{sec:hard_pruning}
We also observe that the model's performance after densification closely matches that of the fully-trained model. The iterations after the densification stage essentially fine-tune the model and can be used to further ``refine'' it after pruning, similar to PUP 3D-GS~\cite{hansontu2024pup} and LightGaussian~\cite{fan2023lightgaussian}. In practice, our Hard Pruning method prunes the model by a constant ratio every $3000$ iterations starting at iteration $15000$. We Hard Prune $30\%$ of Gaussians in each interval, which, when paired with Soft Pruning, empirically reduces the total number of Gaussians across scenes by $10.6\times$.
\section{Experiments}
\label{sec:experiments}

\subsection{Datasets}
\label{sec:datasets}
Our evaluation uses the same set of challenging real-world scenes as 3D-GS~\cite{kerbl3Dgaussians}. This includes nine Mip-Nerf 360 scenes~\cite{barron2022mipnerf360} -- four indoor and five outdoor -- that each feature a complex central object or area with a detailed background. We also include the outdoor \emph{train} and \emph{truck} scenes from the Tanks \& Temples dataset~\cite{Knapitsch2017tandt} and the indoor \emph{drjohnson} and \emph{playroom} scenes from the Deep Blending dataset~\cite{DeepBlending2018}. For consistency across experiments, we use the COLMAP pose estimates and sparse point clouds provided by the dataset authors.

\subsection{Implementation Details}

Our code builds on the differentiable renderer provided by 3D-GS~\cite{kerbl3Dgaussians} and modifies the Python training pipeline for pruning schedules and execution. To ensure consistent and precise timing, all times in Table~\ref{tab:time_table_full} and FPS values in Tables~\ref{tab:our_raw_numbers} and~\ref{tab:main_metric_table} are measured using \texttt{CUDA} events at the start and end of the forward rendering procedure. All experiments are conducted on an Nvidia RTXA5000 GPU, and the reported \acronym{} results represent the average metrics across three independent runs for each scene.

\begin{table}[t]
\caption{Average Gaussian count, FPS, and training time across all scenes in Section~\ref{sec:datasets}. Ratios for model size compression, rendering speed-up, and training speed-up are reported in (parentheses). The operation in each row is applied \textbf{cumulatively} to all of the following rows. The \ctext[RGB]{134,230,85}{best} and \ctext[RGB]{208,240,192}{second} best value for each metric are color coded.}
\centering
\vspace{-1mm}
\resizebox{\columnwidth}{!}{
\begin{tabular}{l|ccc}
\toprule
Method &
\# Gaussians~$\downarrow$ &
FPS~$\uparrow$ &
Training Time~$\downarrow$ \\
\midrule
Baseline &
2.93M &
134 &
23.2 \\
+SnugBox &
2.97M &
244 (1.82$\times$) &
21.2 (1.09$\times$) \\
+AccuTile &
2.97M &
267 (1.99$\times$) &
21.0 (1.10$\times$) \\
+Soft Pruning &
\metrictablesecond{1.64M (1.79$\times$)} & 
\metrictablesecond{420 (3.14$\times$)} & 
\metrictablesecond{17.5 (1.32$\times$)} \\
+Hard Pruning &
\metrictablebest{0.28M (10.6$\times$)} & 
\metrictablebest{898 (6.71$\times$)} &
\metrictablebest{15.7 (1.47$\times$)} \\
\bottomrule
\end{tabular}
}
\label{tab:our_raw_numbers}
\vspace{-3mm}
\end{table}

\begin{table}[t]
\caption{Average reported metrics for each pruning method across all scenes in the Mip-NeRF 360 dataset. The \textit{Comp} column reports model size compression in terms of Gaussian count, \textit{FPS} reports rendering speed-up, and \textit{Train} reports training time speed-up, all with respect to the baseline 3D-GS model. PSNR, SSIM, and LPIPS are also recorded. For a fair comparison, we report the published results of each method and use `-' to denote missing metrics. The \ctext[RGB]{134,230,85}{best} and \ctext[RGB]{208,240,192}{second} best value for each metric are color coded; lossless methods are \underline{underlined}. Results for the Tanks \& Temples and Deep Blending datasets are reported in Appendix~\ref{sec:appendix:dataset_eval}.}
\centering
\resizebox{\columnwidth}{!}{
\begin{tabular}{l|ccc|ccc}
\toprule
{\small Method} & {\small Comp~$\uparrow$} & {\small FPS~$\uparrow$} & {\small Train~$\uparrow$} & {\small PSNR~$\uparrow$} & {\small SSIM~$\uparrow$} & {\small LPIPS~$\downarrow$} \\
\midrule
3D-GS \cite{kerbl3Dgaussians} & 1.00$\times$ & 1.00$\times$ & 1.00$\times$ & 27.55 & 0.814 & 0.222 \\
Trimming \cite{ali2024trimming} & 4.00$\times$ & - & - & 27.13 & 0.798 & 0.248 \\
Compact \cite{lee2023compact} & 2.28$\times$ & 1.07$\times$ & 0.73$\times$ & 27.08 & 0.798 & 0.247 \\
EAGLES \cite{girish2023eagles} & 3.68$\times$ & 1.51$\times$ & \metrictablesecond{1.37$\times$} & 26.94 & 0.800 & 0.250 \\
Reducing \cite{papantonakis2024reducing} & 2.33$\times$ & 1.60$\times$ & 1.23$\times$ & 27.10 & 0.809 & 0.226 \\
Light \cite{fan2023lightgaussian} & 2.94$\times$ & 1.76$\times$ & - & 27.28 & 0.805 & 0.243 \\
ELMGS \cite{ali2024elmgs} & 5.00$\times$ & 2.69$\times$ & - & 27.00 & 0.779 & 0.286 \\
PUP \cite{hansontu2024pup} & \metrictablesecond{8.65$\times$} & 2.55$\times$ & - & 26.83 & 0.792 & 0.268 \\
\underline{Mini-Splat \cite{fang2024mini}} & 6.84$\times$ & \metrictablesecond{3.20$\times$} & 1.26$\times$ & 27.34 & \metrictablebest{0.822} & \metrictablebest{0.217} \\
\cmidrule{1-7}
\underline{+SnugBox} & 0.99$\times$ & 1.81$\times$ & 1.08$\times$ & \metrictablesecond{27.55} & \metrictablesecond{0.814} & \metrictablesecond{0.221} \\
\underline{+AccuTile} & 0.99$\times$ & 1.99$\times$ & 1.10$\times$ & \metrictablebest{27.57} & \metrictablesecond{0.814} & \metrictablesecond{0.221} \\
+Soft Pruning & 1.79$\times$ & 3.14$\times$ & 1.30$\times$ & 27.32 & 0.807 & 0.246 \\
+Hard Pruning & \metrictablebest{10.6$\times$} & \metrictablebest{6.51$\times$} & \metrictablebest{1.45$\times$} & 26.94 & 0.782 & 0.296 \\
\bottomrule
\end{tabular}
}
\label{tab:main_metric_table}
\vspace{-2mm}
\end{table}

\subsection{Results}

\subsubsection{Additive method performance}
The efficacy of Speedy-Splat is demonstrated by Table~\ref{tab:time_table_full}, where we record the average execution times of each function across all scenes when additively applying our methods. SnugBox and AccuTile each introduce minimal additional computation time to \texttt{preprocess} and \texttt{InclusiveSum}. However, limiting the number of tiles touched accelerates all downstream functions, culminating in an overall speed-up of $1.82\times$ by SnugBox that is raised to $1.99\times$ by AccuTile. Applying soft pruning reduces the runtime of all functions by reducing the number of Gaussians, leading to a $3.14\times$ overall speed-up. Finally, performing hard pruning improves overall speed by a whopping $6.71\times$ over the baseline 3D-GS model.

\subsubsection{Overall Performance}

In Figures~\ref{fig:teaser} and~\ref{fig:experiments}, we report qualitative results on two outdoor and two indoor scenes from all three datasets. The magnified regions highlight that our method preserves the fine details in the baseline 3D-GS scene and closely models the ground truth view. Despite touting similar compression ratios and rendering nearly identical images, Speedy-Splat achieves over double the FPS of PUP 3D-GS. 

Table~\ref{tab:main_metric_table} compares our methods against other methods that reduce the number of Gaussians and increase inference speed using the mean of each metric across all scenes in the Mip-NeRF 360 dataset. The underlined methods are ``lossless'', meaning that they avoid degrading visual fidelity at all. SnugBox and AccuTile, our lossless methods, improve rendering and training speed while leaving image quality metrics essentially unchanged or slightly better. Our full pipeline, labeled as ``+Hard Pruning" boasts the highest compression ratios, rendering speeds, and training speed-ups across all datasets. Furthermore, its image quality metrics are competitive with the other methods. 

\subsection{Pruning Score Comparison}
\vspace{-0.5mm}
Although the primary focus of our work is rendering speed, we find that our efficient pruning score, described in Section~\ref{sec:efficient_pruning}, also performs well when applied in other compression pipelines. In Table~\ref{tab:speedy_vs_pup}, we ablate our efficient pruning score with the PUP 3D-GS sensitivity score across all scenes in their post-hoc pruning pipeline. Notably, \acronym{}'s efficient pruning score outperforms PUP 3D-GS on PSNR and is competitive across the other metrics.

\begin{table}[t]
\caption{Average metrics across all scenes in Section~\ref{sec:datasets} when using the \acronym{} and PUP 3D-GS pruning scores to prune $88.44\%$ of Gaussians using the PUP 3D-GS pipeline. Two rounds of prune-refine are performed on each baseline 3D-GS model, pruning $66\%$ of Gaussians and then fine-tuning for $5,000$ iterations in each one.
The \ctext[RGB]{134,230,85}{best} value for each metric is color coded.}
\centering
\resizebox{\columnwidth}{!}{
\begin{tabular}{l|ccccc}
\toprule
Method &
\# Gaussians~$\downarrow$ &
PSNR~$\uparrow$ &
SSIM~$\uparrow$ &
LPIPS~$\downarrow$ &
FPS~$\uparrow$ \\
\midrule
Baseline &
2.92M &
27.1503 &
0.8296 & 
0.2238 &
107.53 \\
PUP~\cite{hansontu2024pup} &
0.34M &
26.2136 &
\metrictablebest{0.8044} & 
\metrictablebest{0.2731} &
\metrictablebest{378.57}
\\
Ours &
0.34M &
\metrictablebest{26.8658} &
0.8022 & 
0.2840 &
345.52 \\
\bottomrule
\end{tabular}
}
\label{tab:speedy_vs_pup}
\vspace{-2.9mm}
\end{table}

\vspace{-0.25mm}
\section{Limitations}
\vspace{-1.25mm}
\label{sec:limitations}
A limitation of \acronym{} is that it produces slightly lower image quality than 3D-GS. However, this degradation is expected at high compression ratios and is also observed in comparable techniques. Additionally, a direct comparison of our efficient pruning score to the PUP 3D-GS pruning score illuminates a slight, yet noticeable, gap in performance. Future work could explore the possibility of another efficient pruning score that delivers higher performance. 

\vspace{-0.25mm}
\section{Conclusion}
\vspace{-1.25mm}

In this work, we present \acronym{}: a new 3D-GS technique that accurately localizes Gaussians during rendering and significantly improves inference speed, model size, and training time. Enhanced localization is achieved by our SnugBox and AccuTile methods, while model size reduction is accomplished by our Soft and Hard Pruning approaches. Together, our \acronym{} methods accelerate rendering speed by an average of $6.71\times$, reduce model size by  $10.6\times$, and improve training time by $1.47\times$ across all scenes from the Mip-NeRF 360, Tanks \& Temples, and Deep Blending datasets.
\vspace{-0.75mm}
\section{Acknowledgements}
\vspace{-1.75mm}
This work was made possible by the IARPA WRIVA Program, the ONR MURI program, and DAPRA TIAMAT. Commercial support was provided by Capital One Bank, the Amazon Research Award program, and Open Philanthropy. Further support was provided by the National Science Foundation (IIS-2212182), and by the NSF TRAILS Institute (2229885). Zwicker was additionally supported by the National Science Foundation (IIS-2126407).

{
    \small
    \bibliographystyle{ieeenat_fullname}
    \bibliography{main}
}

\clearpage
\appendix

\section{Appendix}

\subsection{AccuTile Proof of Correctness Sketch}
\label{sec:appendix:proof}

We outline the correctness of our AccuTile algorithm by examining the different cases that arise when identifying the minimum and maximum points of an ellipse within a given tile row. Due to the symmetry of Equation~\ref{eq:simplified_ellipse}, exchanging the variables $x$ and $y$ along with the coefficients $a$ and $c$ yields equivalent statements for tile columns. Thus, we focus our discussion on tile rows.

\begin{figure}[h]
  \includegraphics[width=\linewidth]{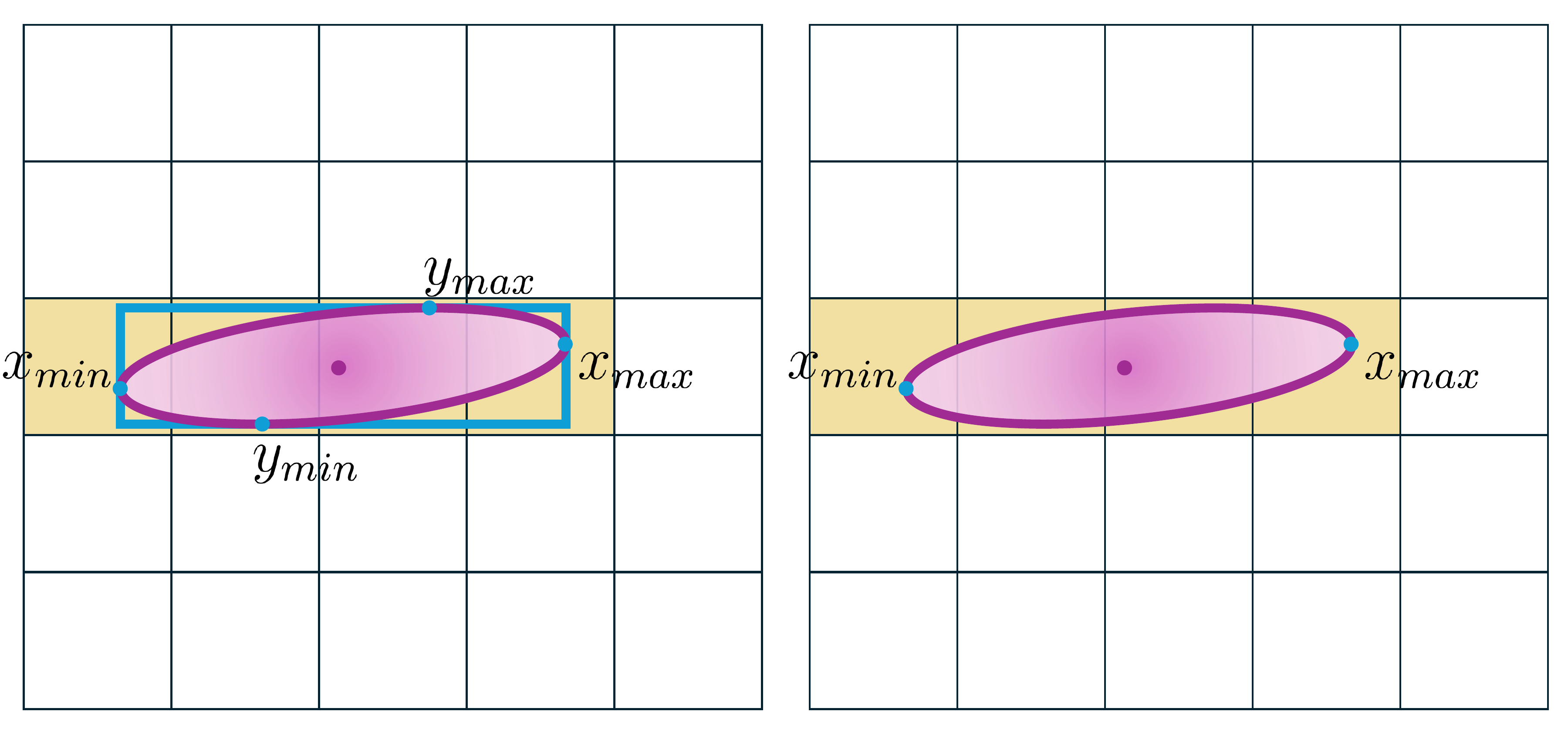}
    \caption{(Left) SnugBox and (right) AccuTile sketch of Case 1. As with Figure~\ref{fig:accutile_tile_intersect}, our AccuTile algorithm iterates over the tile rows; the only points that are processed are $x_{min}$ and $x_{max}$.}
  \label{fig:sketch_a}
\end{figure}

\vspace{0.5em}
\noindent
\textbf{Case 1:} The ellipse does not intersect the tile row boundary.
\noindent
The entire ellipse, including the bounding box extrema $x_{\min}$ and $x_{\max}$ computed by SnugBox, lies within the row. AccuTile correctly selects these points as the furthest ellipse extents. Figure~\ref{fig:sketch_a} illustrates an example of this.

\vspace{0.5em}
\noindent
\textbf{Case 2:} The ellipse intersects one of the tile row bounding lines but not the other.

\begin{figure}[h]
  \includegraphics[width=\linewidth]{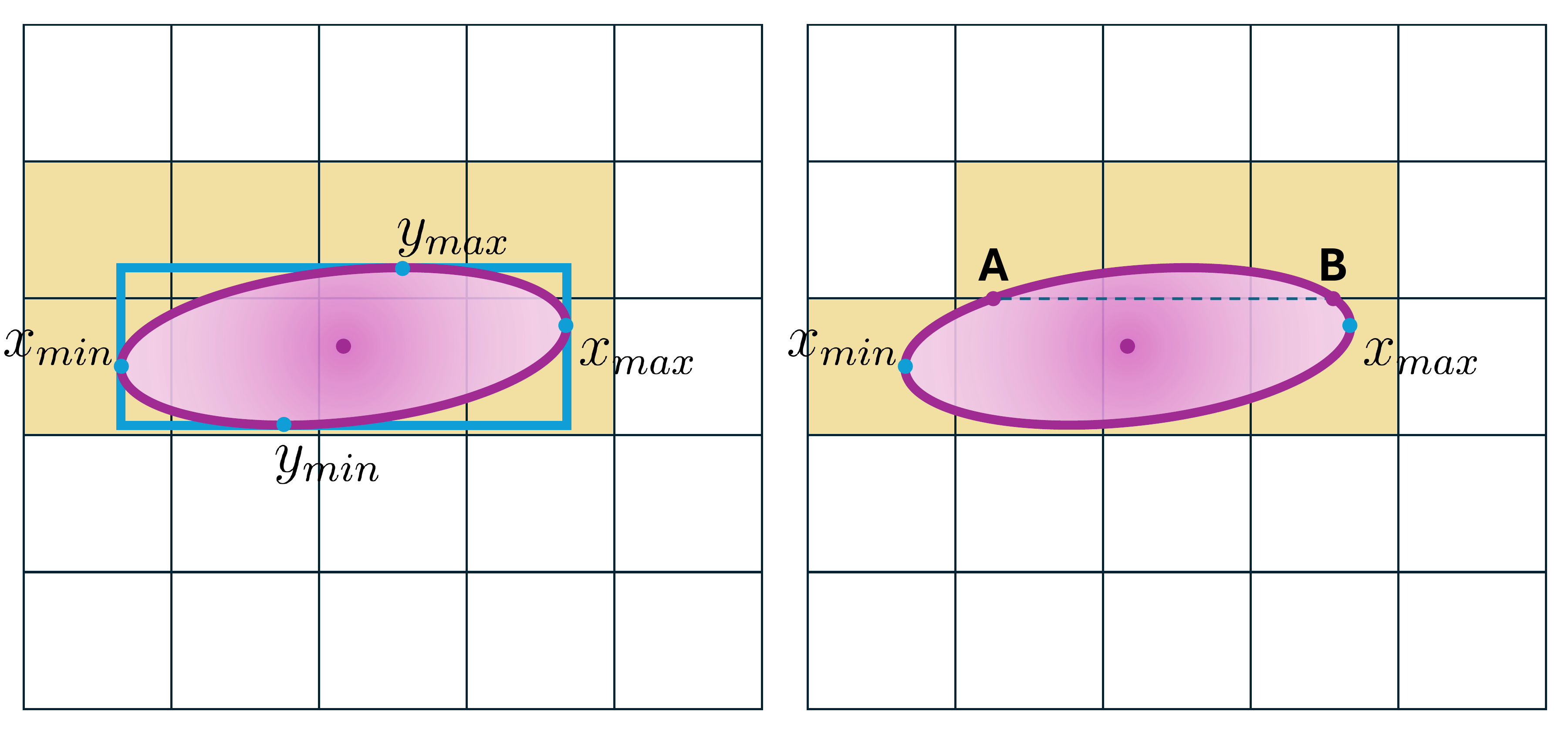}
    \caption{(Left) SnugBox and (right) AccuTile sketch of Case 2.1. Our AccuTile algorithm iterates over the tile rows; the only points that are processed are $x_{min}$, $x_{max}$, \textbf{A}, and \textbf{B}.}
  \label{fig:sketch_b}
\end{figure}

\begin{figure}[h]
  \includegraphics[width=\linewidth]{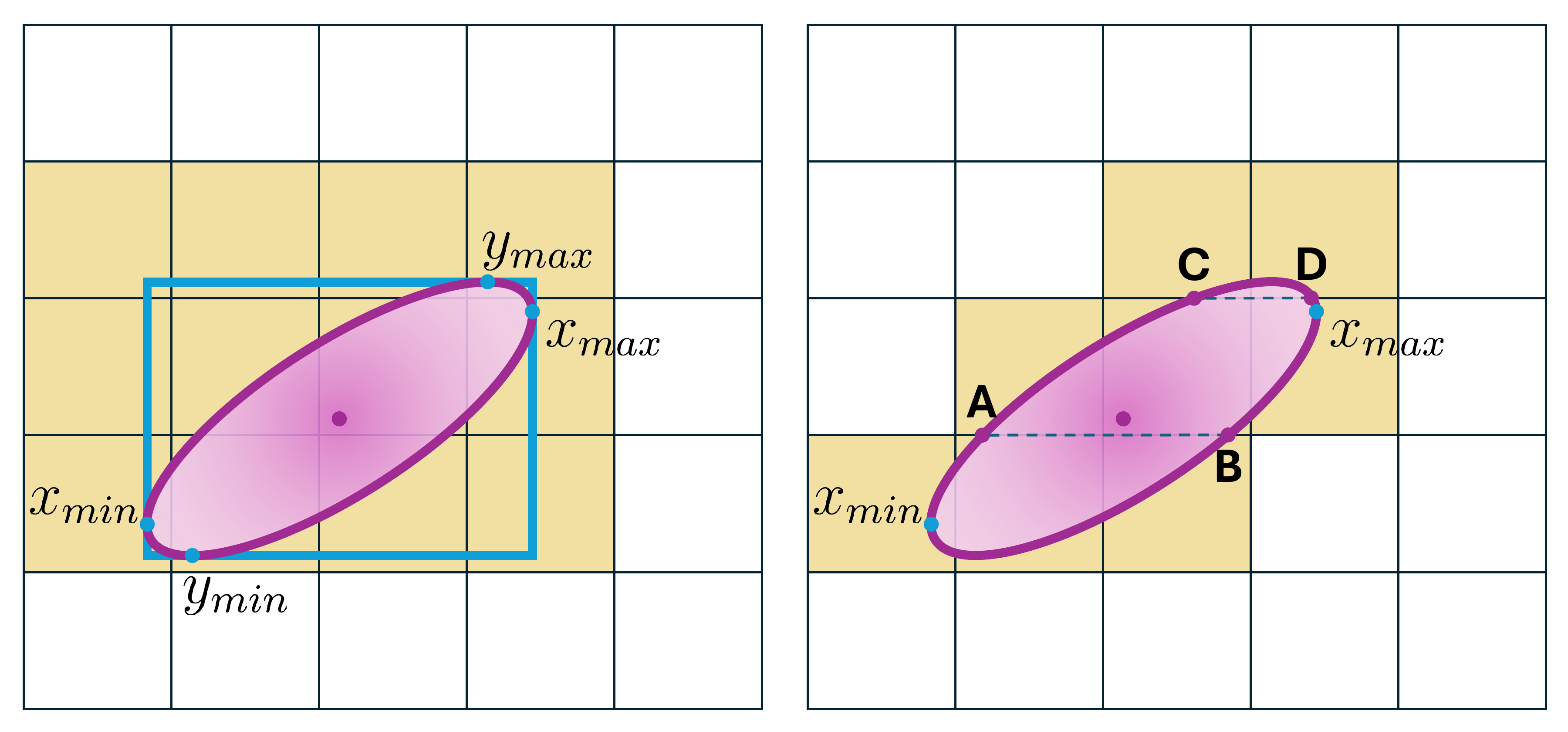}
    \caption{(Left) SnugBox and (right) AccuTile sketch of Cases 2.2, 2.3, and 3. Our AccuTile algorithm iterates over the tile rows; the only points that are processed are \textbf{$x_{min}$}, \textbf{$x_{max}$}, \textbf{A}, \textbf{B}, \textbf{C}, and \textbf{D}. A detailed walkthrough of this example is presented in Section~\ref{sec:accutile}.}
  \label{fig:sketch_c}
\end{figure}

\noindent
This implies that either $y_{\min}$ or $y_{\max}$ lies within the row, but not both. There are several possible subcases:

\begin{itemize}
\item{\textbf{Case 2.1:}} If both $x_{\min}$ and $x_{\max}$ are in the tile row, then they are correctly assigned as the furthest extent of the ellipse by AccuTile. Figure~\ref{fig:sketch_b} illustrates an example of this case.

\item{\textbf{Case 2.2:}} If $x_{\min}$ and $x_{\max}$ are not in the row but $y_{\max}$ is, then the ellipse decreases monotonically from $y_{\max}$ to the row boundary on both sides of $y_{\max}$, making the row boundary intersections the furthest extent of the ellipse and are the points selected by AccuTile as the furthest row extent. This follows from the absence of the critical points $x_{\min}$ and $x_{\max}$. A symmetric argument applies when $y_{\min}$ is in the row instead. The top tile row of Figures~\ref{fig:sketch_b} and \ref{fig:sketch_c} illustrate examples of this case.

\item{\textbf{Case 2.3:}} If $x_{\min}$ and $y_{\min}$ are in the row but $x_{\max}$ is not, then $x_{\min}$ is assigned as the minimum extent of the ellipse. The ellipse increases monotonically from $y_{\min}$ to the boundary to the right of $y_{\min}$ and from $x_{\min}$ to the boundary to the right of $x_{\min}$. Under a corrollary of the definition of an ellipse, the sides of the ellipse do not intersect. Thus, the ellipse point on the curve that extends to the right of $y_{\min}$ and intersects the tile row boundary must be the maximum ellipse extent, and AccuTile correctly selects it as such. A similar argument applies in the following cases: (1) $x_{\max}$ and $y_{\min}$ are in the tile row but $x_{\min}$ is not, (2) $x_{\min}$ and $y_{\max}$ are in the tile row but $x_{\max}$ is not, and (3) $x_{\max}$ and $y_{\max}$ are in the tile row but $x_{\min}$ is not. The bottom tile row of Figure~\ref{fig:sketch_c} illustrates an example of this case.

\end{itemize}

\vspace{0.5em}
\noindent
\textbf{Case 3:} The ellipse intersects both the top and bottom row boundary.

\noindent If $x_{\min}$ or $x_{\max}$ is in the tile row, then AccuTile correctly assigns it as the minimum or maximum extent of the ellipse, respectively. The right side of the ellipse in the middle tile row in Figure~\ref{fig:sketch_c} illustrates an example of this case. Otherwise, the ellipse monotonically increases from the bottom row boundary to the top row boundary, or vice-versa, due to the absence of critical points. Selecting the minimum or maximum boundary point, as done by AccuTile, yields the correct result. The left side of the ellipse in the middle tile row in Figure~\ref{fig:sketch_c} illustrates an example of this case.

\begin{figure*}[t]
  \includegraphics[width=\linewidth]{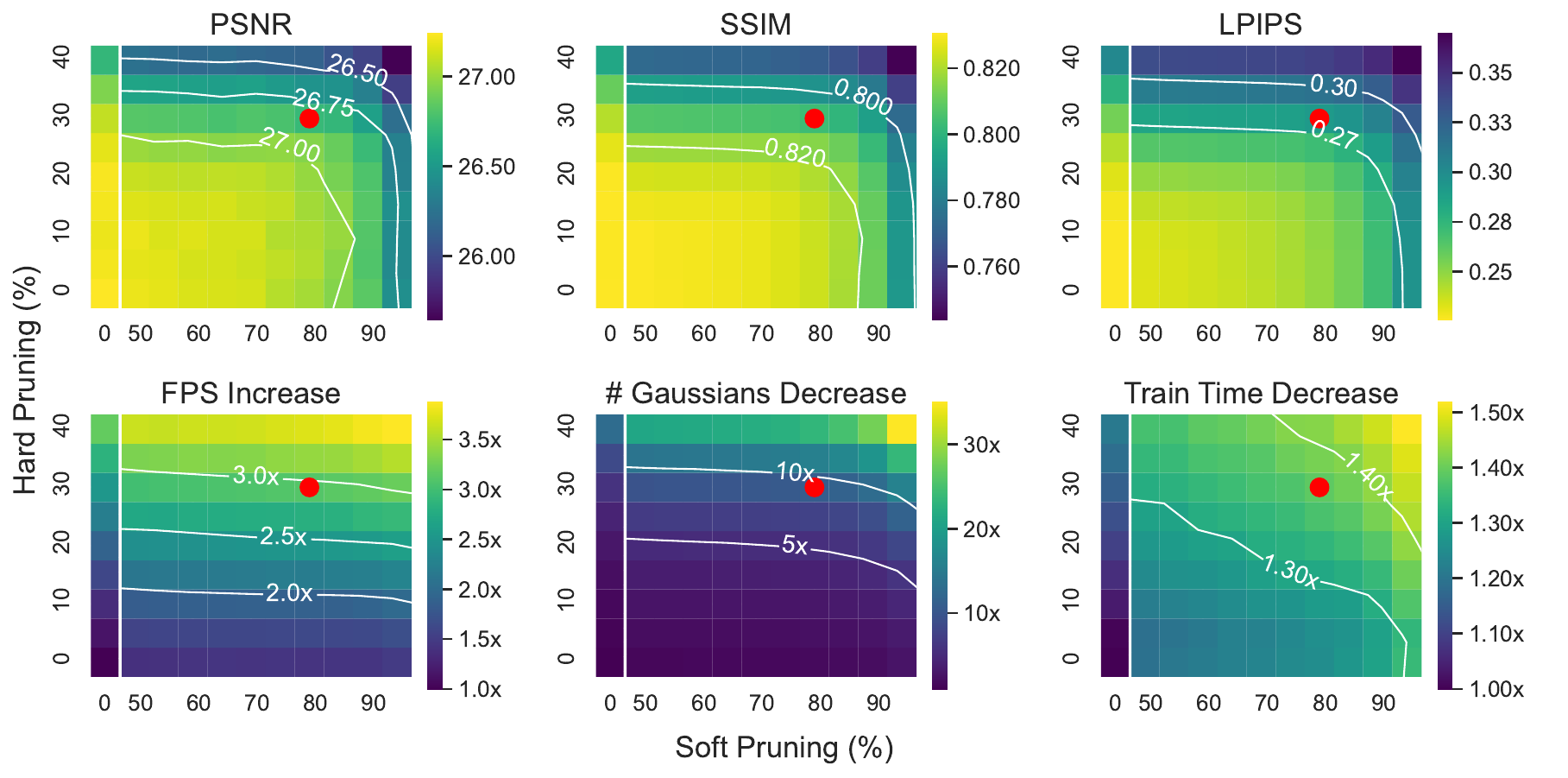}
  \vspace{-7mm}
    \caption{We sweep pruning percentages in $5\%$ increments for Hard Pruning ($0\%-40\%$) and Soft Pruning ($0\%$, $50\%-95\%$) on all scenes listed in Section~\ref{sec:datasets}. Experiments are performed $3\times$ on each scene without our Gaussian localization methods; the reported metrics are averaged across all runs. ($0\%$,~$0\%$) is the baseline 3D-GS model, the first column ($0\%$,~:) is Hard Pruning in isolation, and the first row (:,~$0\%$) is Soft Pruning in isolation. The red dots at ($80\%$, $30\%$) denote the percentage settings used in our manuscript. We report the FPS increase and the Number of Gaussians and Train Time decrease factors to be consistent with the format in Table~\ref{tab:main_metric_table}.}
  \label{fig:all_pruning}
\end{figure*}
\newpage

\begin{table}[h]
\caption{Average reported metrics for each pruning method across all scenes in the Tanks \& Temples dataset.}
\centering
\resizebox{\columnwidth}{!}{
\begin{tabular}{l|ccc|ccc}
\toprule
{\small Method} & {\small Comp~$\uparrow$} & {\small FPS~$\uparrow$} & {\small Train~$\uparrow$} & {\small PSNR~$\uparrow$} & {\small SSIM~$\uparrow$} & {\small LPIPS~$\downarrow$} \\
\midrule
3D-GS \cite{kerbl3Dgaussians} & 1.00$\times$ & 1.00$\times$ & 1.00$\times$ & 23.70 & 0.849 & 0.178 \\
Trimming \cite{ali2024trimming} & 4.00$\times$ & - & - & 23.69 & 0.831 & 0.210 \\
Compact \cite{lee2023compact} & 2.19$\times$ & 1.16$\times$ & 0.76$\times$ & 23.32 & 0.831 & 0.201 \\
EAGLES \cite{girish2023eagles} & - & 1.73$\times$ & 1.19$\times$ & 23.10 & 0.820 & 0.220 \\
Reducing \cite{papantonakis2024reducing} & 2.56$\times$ & 1.91$\times$ & 1.27$\times$ & 23.57 & 0.840 & 0.188 \\
Light \cite{fan2023lightgaussian} & 2.94$\times$ & 1.97$\times$ & - & 23.11 & 0.817 & 0.231 \\
ELMGS \cite{ali2024elmgs} & 5.00$\times$ & \metrictablesecond{4.05$\times$} & - & \metrictablebest{23.90} & 0.825 & 0.233 \\
PUP \cite{hansontu2024pup} & \metrictablesecond{10.0$\times$} & 4.00$\times$ & - & 22.72 & 0.801 & 0.244 \\
\underline{Mini-Splat \cite{fang2024mini}} & 9.20$\times$ & - & - & 23.18 & 0.835 & 0.202 \\
\cmidrule{1-7}
\underline{+SnugBox} & 0.99$\times$ & 1.61$\times$ & 1.11$\times$ & 23.69 & \metrictablebest{0.849} & \metrictablesecond{0.178} \\
\underline{+AccuTile} & 0.99$\times$ & 1.67$\times$ & 1.12$\times$ & \metrictablesecond{23.73} & \metrictablebest{0.849} & \metrictablebest{0.177} \\
+Soft Pruning & 1.69$\times$ & 2.48$\times$ & \metrictablesecond{1.36$\times$} & 23.54 & 0.841 & 0.201 \\
+Hard Pruning & \metrictablebest{10.1$\times$} & \metrictablebest{6.30$\times$} & \metrictablebest{1.58$\times$} & 23.45 & 0.821 & 0.241 \\
\bottomrule
\end{tabular}
}
\label{tab:main_metric_tnt}
\vspace{-2mm}
\end{table}

\subsection{Overall Pruning Percent Metrics}

In Figure~\ref{fig:all_pruning}, we perform a parameter sweep over Hard Pruning percentages from $0\%-40\%$ at $5\%$ intervals and Soft Pruning percentages at $0\%$ and from $50-95\%$ at $5\%$ intervals. We conduct each experiment $3\times$ on each scene listed in Section~\ref{sec:datasets} to reduce variance, then average the metrics across all runs. All experiments are run without our Gaussian localization methods -- SnugBox and AccuTile -- to ablate the effect of each pruning method in isolation. Our ($80\%$, $30\%$) pruning percentages are empirically selected to produce a favorable balance between speed and quality. 

\newpage
\begin{table}[h]
\caption{Average reported metrics for each pruning method across all scenes in the Deep Blending dataset.}
\centering
\resizebox{\columnwidth}{!}{
\begin{tabular}{l|ccc|ccc}
\toprule
{\small Method} & {\small Comp~$\uparrow$} & {\small FPS~$\uparrow$} & {\small Train~$\uparrow$} & {\small PSNR~$\uparrow$} & {\small SSIM~$\uparrow$} & {\small LPIPS~$\downarrow$} \\
\midrule
3D-GS \cite{kerbl3Dgaussians} & 1.00$\times$ & 1.00$\times$ & 1.00$\times$ & 29.09 & 0.886 & 0.288 \\
Trimming \cite{ali2024trimming} & 1.33$\times$ & - & - & 29.43 & 0.897 & 0.267 \\
Compact \cite{lee2023compact} & 2.65$\times$ & 1.37$\times$ & 0.79$\times$ & 29.79 & 0.901 & 0.258 \\
EAGLES \cite{girish2023eagles} & - & 1.30$\times$ & 1.31$\times$ & \metrictablesecond{29.92} & 0.900 & \metrictablesecond{0.250} \\
Reducing \cite{papantonakis2024reducing} & 2.86$\times$ & 1.79$\times$ & 1.27$\times$ & 29.63 & \metrictablesecond{0.902} & \metrictablebest{0.249} \\
Light \cite{fan2023lightgaussian} & - & - & - & - & - & - \\
ELMGS \cite{ali2024elmgs} & 5.00$\times$ & 4.15$\times$ & - & 29.24 & 0.894 & 0.273 \\
PUP \cite{hansontu2024pup} & \metrictablesecond{10.0$\times$} & \metrictablesecond{4.51$\times$} & - & 28.85 & 0.881 & 0.301 \\
\underline{Mini-Splat \cite{fang2024mini}} & 8.06$\times$ & - & - & \metrictablebest{29.98} & \metrictablebest{0.908} & 0.253 \\
\cmidrule{1-7}
\underline{+SnugBox} & 0.97$\times$ & 2.11$\times$ & 1.12$\times$ & 29.18 & 0.886 & 0.287 \\
\underline{+AccuTile} & 0.97$\times$ & 2.32$\times$ & 1.13$\times$ & 29.12 & 0.885 & 0.288 \\
+Soft Pruning & 1.86$\times$ & 3.56$\times$ & \metrictablesecond{1.41$\times$} & 29.29 & 0.889 & 0.296 \\
+Hard Pruning & \metrictablebest{11.1$\times$} & \metrictablebest{7.46$\times$} & \metrictablebest{1.57$\times$} & 29.32 & 0.887 & 0.311 \\
\bottomrule
\end{tabular}

}
\label{tab:main_metric_db}
\vspace{-2mm}
\end{table}

\subsection{Additional Datasets Evaluation}
\label{sec:appendix:dataset_eval}
Table~\ref{tab:main_metric_tnt} and Table~\ref{tab:main_metric_db} present the average reported metrics for each pruning method across all scenes in the Tanks \& Temples and Deep Blending datasets, respectively. The \textit{Comp} column reports model size compression in terms of Gaussian count, \textit{FPS} reports rendering speed-up, and \textit{Train} reports training time speed-up, all with respect to the baseline 3D-GS model. PSNR, SSIM, and LPIPS are also recorded. The \ctext[RGB]{134,230,85}{best} and \ctext[RGB]{208,240,192}{second} best value for each metric are color coded; lossless methods are \underline{underlined}.

\begin{table*}[h]
\caption{
Average execution time (milliseconds) of each function across all scenes. This experiment ablates the StopThePop~\cite{radl2024stopthepop} Tile-Based Culling method with warp-level load balancing against our AccuTile algorithm. For each method, execution times are averaged over three runs, with each run rendering the test set 20 times to reduce variance.  The \ctext[RGB]{134,230,85}{fastest} times are highlighted. Our AccuTile algorithm outperforms the Tile-Based Culling method in overall runtime by a notable margin. For detailed analysis, see Section~\ref{sec:stp-ablation}.}
\centering
\resizebox{\textwidth}{!}{
\begin{tabular}{l|cccccc|c}
\toprule
Method &
Preprocess &
Inclusive Sum &
Duplicate with Keys &
Radix Sort &
Identify Tile Ranges &
Render &
Overall \\
\midrule
Baseline &
0.665 &
0.046 &
0.568 &
1.551 &
0.083 &
4.469 &
7.457
\\
Tile-Based Culling~\cite{radl2024stopthepop} &
0.811 (0.820x) &
0.046 &
0.450 (1.263x) &
\timetablefastest{0.609 (2.548x)} &
\timetablefastest{0.035 (2.341x)} &
\timetablefastest{2.027 (2.205x)} &
4.051 (1.841x)
\\
AccuTile (Ours) &
\timetablefastest{0.659} &
0.046 &
\timetablefastest{0.194 (2.931x)} &
0.610 (2.541x) &
0.035 (2.338x) &
2.042 (2.189x) &
\timetablefastest{3.660 (2.038x)}
\\
\bottomrule
\end{tabular}
}
\label{tab:time_table_stp_method_ablation}
\end{table*}

\begin{table*}[t]
\caption{
Average execution time (milliseconds) of each scene. This experiment ablates the StopThePop~\cite{radl2024stopthepop} Tile-Based Culling method with warp-level load balancing against our AccuTile algorithm by breaking out the per-scene execution times, which were averaged in the Overall column of Table~\ref{tab:time_table_stp_method_ablation}. The \ctext[RGB]{134,230,85}{fastest} times are highlighted. Our AccuTile algorithm outperforms Tile-Based Culling on all scenes.}
\centering
\resizebox{\textwidth}{!}{
\begin{tabular}{l|ccccccccc|cc|cc}
\toprule
 & \multicolumn{9}{c|}{{\small Mip-NeRF 360}} & \multicolumn{2}{c|}{{\small Tanks \& Temples}} & \multicolumn{2}{c}{{\small Deep Blending}} \\
{\small Method} & {\small bicycle} & {\small bonsai} & {\small counter} & {\small flowers} & {\small garden} & {\small kitchen} & {\small room} & {\small stump} & {\small treehill} & {\small train} & {\small truck} & {\small drjohnson} & {\small playroom} \\
\midrule
Baseline & 14.034 & 4.977 & 7.025 & 7.914 & 6.103 & 11.025 & 8.562 & 5.776 & 7.088 & 6.993 & 4.872 & 7.253 & 5.322 \\
Tile-Based Culling~\cite{radl2024stopthepop} & 6.609 & 2.674 & 3.261 & 3.474 & 3.888 & 6.994 & 4.741 & 2.997 & 3.058 & 4.417 & 2.860 & 4.131 & 3.554 \\
AccuTile (Ours) &
\timetablefastest{5.880} &
\timetablefastest{2.401} &
\timetablefastest{2.989} &
\timetablefastest{2.933} &
\timetablefastest{3.478} &
\timetablefastest{6.433} &
\timetablefastest{4.490} &
\timetablefastest{2.578} &
\timetablefastest{2.726} &
\timetablefastest{3.965} &
\timetablefastest{2.729} &
\timetablefastest{3.671} &
\timetablefastest{3.300} \\
\bottomrule
\end{tabular}
}
\label{tab:time_table_stp_scene_ablation}
\end{table*}


\FloatBarrier

\subsection{StopThePop Tile-Based Culling Ablation}
\label{sec:stp-ablation}

We ablate our AccuTile algorithm against the StopThePop~\cite{radl2024stopthepop} Tile-Based Culling method in Tables~\ref{tab:time_table_stp_method_ablation} and~\ref{tab:time_table_stp_scene_ablation}. Tile-Based Culling computes a precise Gaussian-to-Tile mapping in two steps: (1) Similar to our SnugBox method, a tight, opacity-aware bounding box is computed per Gaussian; however, due to the use of thresholds in their code, not all bounding boxes are tight. (2) Each tile touching the bounding box is iteratively examined to determine if it should be included in the final Gaussian-to-Tile mapping; warp-level load balancing is used to accelerate this process.

For this ablation, we update the 3D-GS rasterizer with the Tile-Based Culling code to isolate its runtime speed-up. All warp-level load balancing code is included to ensure that we compare against the most optimized version of the method. As noted in the StopThePop codebase, a padded alpha threshold is required to accurately compute bounding boxes, which, by extension, prevents undercounting Gaussian-to-Tile mappings. No padded alpha threshold is provided, so we perform this ablation without it. To ensure a fair comparison, we train three models for each scene and measure execution times with the baseline 3D-GS, Tile-Based Culling, and AccuTile renderers on each one.

Table~\ref{tab:time_table_stp_method_ablation} shows that our AccuTile method significantly outperforms Tile-Based Culling on Preprocess and Duplicate with Keys. Since Tile-Based Culling iterates over all candidate tiles while AccuTile does not, it requires more computation and induces a markedly higher runtime cost even with warp-level load-balancing. Surprisingly, Tile-Based Culling slightly outperforms AccuTile in the downstream functions Radix Sort, Identify Tile Ranges, and Render. However, we observe that this is caused by the aforementioned under-counting of Gaussian-to-Tile mappings; this marginal improvement disappears when padded alpha thresholds are introduced, further slowing down Preprocess and Duplicate with Keys. Additionally, as reported by Table~\ref{tab:time_table_stp_scene_ablation}, our AccuTile method consistently outperforms Tile-Based Culling across all scenes.

\subsection{Per-Scene Metrics}
PSNR, SSIM, LPIPS, FPS, and training times for each scene from the Mip-NeRF 360, Tanks\&Temples, and Deep Blending datasets that was used in 3D-GS~\cite{kerbl3Dgaussians} are recorded in Tables~\ref{tab:breakout_psnr_table}, \ref{tab:breakout_ssim_table}, \ref{tab:breakout_lpips_table}, ~\ref{tab:breakout_fps_table}, and Table~\ref{tab:breakout_train_time_table}, respectively. The operation in each row is applied \textbf{cumulatively} to all of the following rows.

\begin{table*}
\caption{PSNR~$\uparrow$ on each scene after cumulatively applying each function.}
\vspace{-2mm}
\centering
\resizebox{\linewidth}{!}{
\begin{tabular}{l|ccccccccc|cc|cc}
\toprule
 & \multicolumn{9}{c|}{{\small Mip-NeRF 360}} & \multicolumn{2}{c|}{{\small Tanks \& Temples}} & \multicolumn{2}{c}{{\small Deep Blending}} \\
{\small Method} & {\small bicycle} & {\small bonsai} & {\small counter} & {\small flowers} & {\small garden} & {\small kitchen} & {\small room} & {\small stump} & {\small treehill} & {\small train} & {\small truck} & {\small drjohnson} & {\small playroom} \\
\midrule
Baseline & 25.10 & 32.42 & 29.14 & 21.41 & 27.31 & 31.49 & 31.66 & 26.78 & 22.62 & 22.01 & 25.40 & 28.18 & 30.00 \\
\midrule
+SnugBox & 25.12 & 32.36 & 29.09 & 21.45 & 27.31 & 31.61 & 31.70 & 26.78 & 22.54 & 21.97 & 25.41 & 28.27 & 30.09 \\
+AccuTile & 25.13 & 32.42 & 29.13 & 21.43 & 27.33 & 31.65 & 31.68 & 26.80 & 22.58 & 22.00 & 25.45 & 28.23 & 30.00 \\
+Soft Pruning & 25.09 & 31.91 & 28.74 & 21.35 & 27.16 & 30.83 & 31.32 & 26.88 & 22.57 & 21.74 & 25.34 & 28.44 & 30.14 \\
+Hard Pruning & 24.78 & 31.29 & 28.28 & 21.21 & 26.70 & 29.91 & 30.99 & 26.79 & 22.51 & 21.71 & 25.20 & 28.50 & 30.14 \\
\bottomrule
\end{tabular}
}
\label{tab:breakout_psnr_table}
\end{table*}

\begin{table*}
\caption{SSIM~$\uparrow$ on each scene after cumulatively applying each function.}
\vspace{-2mm}
\centering
\resizebox{\linewidth}{!}{
\begin{tabular}{l|ccccccccc|cc|cc}
\toprule
 & \multicolumn{9}{c|}{{\small Mip-NeRF 360}} & \multicolumn{2}{c|}{{\small Tanks \& Temples}} & \multicolumn{2}{c}{{\small Deep Blending}} \\
{\small Method} & {\small bicycle} & {\small bonsai} & {\small counter} & {\small flowers} & {\small garden} & {\small kitchen} & {\small room} & {\small stump} & {\small treehill} & {\small train} & {\small truck} & {\small drjohnson} & {\small playroom} \\
\midrule
Baseline & 0.747 & 0.948 & 0.916 & 0.589 & 0.857 & 0.933 & 0.927 & 0.770 & 0.636 & 0.815 & 0.883 & 0.880 & 0.891 \\
\midrule
+SnugBox & 0.749 & 0.948 & 0.916 & 0.591 & 0.857 & 0.933 & 0.928 & 0.771 & 0.636 & 0.815 & 0.883 & 0.880 & 0.892 \\
+AccuTile & 0.749 & 0.948 & 0.916 & 0.590 & 0.857 & 0.933 & 0.927 & 0.771 & 0.637 & 0.816 & 0.883 & 0.879 & 0.891 \\
+Soft Pruning & 0.741 & 0.941 & 0.904 & 0.582 & 0.848 & 0.921 & 0.920 & 0.776 & 0.630 & 0.803 & 0.878 & 0.884 & 0.893 \\
+Hard Pruning & 0.704 & 0.927 & 0.878 & 0.561 & 0.815 & 0.894 & 0.905 & 0.765 & 0.590 & 0.773 & 0.868 & 0.882 & 0.892 \\
\bottomrule
\end{tabular}
}
\label{tab:breakout_ssim_table}
\end{table*}

\begin{table*}
\caption{LPIPS~$\downarrow$ on each scene after cumulatively applying each function.}
\vspace{-2mm}
\centering
\resizebox{\linewidth}{!}{
\begin{tabular}{l|ccccccccc|cc|cc}
\toprule
 & \multicolumn{9}{c|}{{\small Mip-NeRF 360}} & \multicolumn{2}{c|}{{\small Tanks \& Temples}} & \multicolumn{2}{c}{{\small Deep Blending}} \\
{\small Method} & {\small bicycle} & {\small bonsai} & {\small counter} & {\small flowers} & {\small garden} & {\small kitchen} & {\small room} & {\small stump} & {\small treehill} & {\small train} & {\small truck} & {\small drjohnson} & {\small playroom} \\
\midrule
Baseline & 0.244 & 0.183 & 0.185 & 0.359 & 0.122 & 0.118 & 0.200 & 0.242 & 0.346 & 0.208 & 0.147 & 0.291 & 0.284 \\
\midrule
+SnugBox & 0.241 & 0.183 & 0.185 & 0.358 & 0.122 & 0.117 & 0.199 & 0.241 & 0.345 & 0.208 & 0.147 & 0.291 & 0.284 \\
+AccuTile & 0.242 & 0.183 & 0.185 & 0.359 & 0.122 & 0.117 & 0.199 & 0.241 & 0.344 & 0.207 & 0.147 & 0.292 & 0.284 \\
+Soft Pruning & 0.271 & 0.197 & 0.212 & 0.379 & 0.147 & 0.141 & 0.222 & 0.258 & 0.390 & 0.237 & 0.165 & 0.297 & 0.295 \\
+Hard Pruning & 0.333 & 0.231 & 0.260 & 0.419 & 0.213 & 0.198 & 0.260 & 0.288 & 0.463 & 0.291 & 0.191 & 0.313 & 0.308 \\
\bottomrule
\end{tabular}
}
\label{tab:breakout_lpips_table}
\end{table*}

\begin{table*}
\caption{FPS~$\uparrow$ on each scene after cumulatively applying each function. Speed-ups~$\uparrow$ are recorded in (parentheses).}
\vspace{-2mm}
\centering
\resizebox{\linewidth}{!}{
\begin{tabular}{l|ccccccccc|cc|cc}
\toprule
 & \multicolumn{9}{c|}{{\small Mip-NeRF 360}} & \multicolumn{2}{c|}{{\small Tanks \& Temples}} & \multicolumn{2}{c}{{\small Deep Blending}} \\
{\small Method} & {\small bicycle} & {\small bonsai} & {\small counter} & {\small flowers} & {\small garden} & {\small kitchen} & {\small room} & {\small stump} & {\small treehill} & {\small train} & {\small truck} & {\small drjohnson} & {\small playroom} \\
\midrule
Baseline & 71 & 201 & 142 & 164 & 91 & 117 & 140 & 141 & 138 & 200 & 185 & 126 & 172 \\
\midrule
\multirow{2}{*}{+SnugBox} & 154 & 358 & 276 & 267 & 146 & 197 & 301 & 228 & 247 & 320 & 282 & 301 & 335 \\ 
& (2.15$\times$) & (1.78$\times$) & (1.95$\times$) & (1.62$\times$) & (1.60$\times$) & (1.68$\times$) & (2.15$\times$) & (1.61$\times$) & (1.79$\times$) & (1.60$\times$) & (1.53$\times$) & (2.39$\times$) & (1.95$\times$) \\ 
&&&&&&&&&&&& \\[-0.5em]
\multirow{2}{*}{+AccuTile} & 168 & 413 & 330 & 285 & 155 & 221 & 315 & 248 & 272 & 343 & 294 & 332 & 378 \\ 
& (2.35$\times$) & (2.05$\times$) & (2.33$\times$) & (1.73$\times$) & (1.70$\times$) & (1.89$\times$) & (2.25$\times$) & (1.75$\times$) & (1.97$\times$) & (1.71$\times$) & (1.59$\times$) & (2.64$\times$) & (2.20$\times$) \\ 
&&&&&&&&&&&& \\[-0.5em]
\multirow{2}{*}{+Soft Pruning} & 241 & 601 & 505 & 419 & 255 & 425 & 549 & 379 & 423 & 518 & 477 & 497 & 612 \\ 
& (3.37$\times$) & (2.99$\times$) & (3.56$\times$) & (2.55$\times$) & (2.80$\times$) & (3.63$\times$) & (3.92$\times$) & (2.68$\times$) & (3.06$\times$) & (2.59$\times$) & (2.58$\times$) & (3.95$\times$) & (3.56$\times$) \\ 
&&&&&&&&&&&& \\[-0.5em]
\multirow{2}{*}{+Hard Pruning} & 662 & 978 & 842 & 825 & 640 & 809 & 942 & 724 & 957 & 1392 & 1149 & 1122 & 1277 \\ 
& (9.25$\times$) & (4.87$\times$) & (5.94$\times$) & (5.02$\times$) & (7.03$\times$) & (6.90$\times$) & (6.73$\times$) & (5.12$\times$) & (6.93$\times$) & (6.95$\times$) & (6.21$\times$) & (8.91$\times$) & (7.42$\times$) \\ 
\bottomrule
\end{tabular}
}
\label{tab:breakout_fps_table}
\end{table*}

\begin{table*}
\caption{Training time~$\downarrow$ in minutes on each scene after cumulatively applying each function. Speed-ups~$\uparrow$ are recorded in (parentheses).}
\vspace{-2mm}
\centering
\resizebox{\linewidth}{!}{
\begin{tabular}{l|ccccccccc|cc|cc}
\toprule
 & \multicolumn{9}{c|}{{\small Mip-NeRF 360}} & \multicolumn{2}{c|}{{\small Tanks \& Temples}} & \multicolumn{2}{c}{{\small Deep Blending}} \\
{\small Method} & {\small bicycle} & {\small bonsai} & {\small counter} & {\small flowers} & {\small garden} & {\small kitchen} & {\small room} & {\small stump} & {\small treehill} & {\small train} & {\small truck} & {\small drjohnson} & {\small playroom} \\
\midrule
Baseline & 31.9 & 20.4 & 24.1 & 24.1 & 32.3 & 27.8 & 23.7 & 24.1 & 24.2 & 11.1 & 13.4 & 24.8 & 19.5 \\
\midrule
\multirow{2}{*}{+SnugBox} & 28.2 & 19.2 & 21.8 & 22.7 & 29.9 & 25.8 & 21.4 & 22.9 & 22.4 & 9.8 & 12.3 & 21.7 & 17.8 \\
 & (1.13$\times$) & (1.07$\times$) & (1.11$\times$) & (1.06$\times$) & (1.08$\times$) & (1.08$\times$) & (1.11$\times$) & (1.05$\times$) & (1.08$\times$) & (1.13$\times$) & (1.09$\times$) & (1.14$\times$) & (1.09$\times$) \\
 &&&&&&&&&&&& \\[-0.5em]
\multirow{2}{*}{+AccuTile} & 27.8 & 19.0 & 21.3 & 22.6 & 29.4 & 25.5 & 21.1 & 22.7 & 22.3 & 9.7 & 12.2 & 21.5 & 17.7 \\
 & (1.15$\times$) & (1.08$\times$) & (1.13$\times$) & (1.07$\times$) & (1.10$\times$) & (1.09$\times$) & (1.12$\times$) & (1.06$\times$) & (1.08$\times$) & (1.14$\times$) & (1.09$\times$) & (1.15$\times$) & (1.10$\times$) \\
 &&&&&&&&&&&& \\[-0.5em]
\multirow{2}{*}{+Soft Pruning} & 23.1 & 17.0 & 18.6 & 19.5 & 23.2 & 20.3 & 18.3 & 19.3 & 18.9 & 8.3 & 9.7 & 17.3 & 14.2 \\
 & (1.38$\times$) & (1.20$\times$) & (1.30$\times$) & (1.23$\times$) & (1.39$\times$) & (1.37$\times$) & (1.30$\times$) & (1.25$\times$) & (1.27$\times$) & (1.33$\times$) & (1.38$\times$) & (1.43$\times$) & (1.37$\times$) \\
 &&&&&&&&&&&& \\[-0.5em]
\multirow{2}{*}{+Hard Pruning} & 19.7 & 16.0 & 17.7 & 17.5 & 20.3 & 18.7 & 16.9 & 17.1 & 16.9 & 7.2 & 8.3 & 15.3 & 12.8 \\
 & (1.62$\times$) & (1.28$\times$) & (1.36$\times$) & (1.38$\times$) & (1.59$\times$) & (1.49$\times$) & (1.40$\times$) & (1.41$\times$) & (1.43$\times$) & (1.55$\times$) & (1.61$\times$) & (1.62$\times$) & (1.52$\times$) \\
\bottomrule
\end{tabular}
}
\label{tab:breakout_train_time_table}
\end{table*}

\clearpage

\end{document}